\documentclass[conference]{IEEEtran}
\IEEEoverridecommandlockouts

\usepackage[algo2e]{algorithm2e} %
\usepackage[utf8]{inputenc}
\usepackage[english]{babel}
\usepackage{flafter}
\usepackage{microtype}
\usepackage{cite}
\usepackage{graphicx}
\usepackage{subfig}
\usepackage[figurename=Fig.]{caption}
\usepackage{amssymb}
\usepackage{amsmath} %
\usepackage{amsfonts} %
\usepackage[hang,flushmargin]{footmisc}
\usepackage{xcolor} 
\usepackage{hyperref}

\usepackage{placeins}
\usepackage{float}
\usepackage{algorithm}
\usepackage{algorithmic}

\SetKwComment{Comment}{/* }{ */}

\DeclareMathOperator*{\argmin}{arg\,min}

\title{Enhancing AUV Autonomy With Model Predictive Path Integral Control
\thanks{This work was supported by the EPSRC Centre for Doctoral Training in Robotics and Autonomous
Systems, funded by the UK Engineering and Physical Sciences Research Council, the Edinburgh
Centre for Robotics and Vaarst (\url{https://vaarst.com/})}}

\author{Pierre Nicolay\textsuperscript{a}, Yvan Petillot\textsuperscript{b}, Mykhaylo Marfeychuk\textsuperscript{c}, Sen Wang\textsuperscript{d}, Ignacio Carlucho\textsuperscript{b} \\
pon1@hw.ac.uk, y.r.petillot@hw.ac.uk, sen.wang@imperial.ac.uk, ignacio.carlucho@hw.ac.uk \\
\textsuperscript{a }\textit{School of Mathematics \& Computer Sciences,} \textit{Heriot-Watt University}, Edinburgh, UK \\
\textsuperscript{b }\textit{School of Engineering \& Physical Sciences,} \textit{Heriot-Watt University}, Edinburgh, UK \\
\textsuperscript{c }\textit{Instituto Superior T\'ecnico,} \textit{Universidade de Lisboa}, Lisboa, Portugal \\ 
\textsuperscript{d }\textit{Department of Electrical and Electronic Engineering,} \textit{Imperial College}, London, UK \\ 
}

\begin{document}

\maketitle
 \thispagestyle{empty}
\pagestyle{empty}

\begin{abstract}
    Autonomous underwater vehicles (AUVs) play a crucial role in surveying marine environments, carrying out underwater inspection tasks, and ocean exploration. However, in order to ensure that the AUV is able to carry out its mission successfully, a control system capable of adapting to changing environmental conditions is required. Furthermore, to ensure the safe operation of the robotic platform the onboard controller should be able to operate under certain constraints.
    In this work, we investigate the feasibility of Model Predictive Path Integral Control (MPPI) for the control of an AUV. 
   We utilise a non-linear model of the AUV to propagate the samples of the MPPI, which allow us to compute the control action in real-time. We provide a detailed evaluation of the effect of the main hyperparameters on the performance of the MPPI controller.
    Furthermore, we compared the performance of the proposed method with a classical PID and Cascade PID approach, demonstrating the superiority of our proposed controller.
    Finally, we present results where environmental constraints are added and show how MPPI can handle them by simply incorporating those constraints in the cost function.
\end{abstract}


\section{INTRODUCTION}
\label{sec:intro}

Designing control systems for underwater vehicles is challenging for multiple reasons~\cite{Bluerov2}.
First and foremost, the dynamic nature of underwater environments is extremely non-linear and has highly coupled effects~\cite{fossen2002}.
Furthermore, the inaccessibility of the environment, as well as the prevalence of disturbances such as tide currents, vortexes and waves, make controlling underwater vehicles very complex. Additionally,
the low bandwidth acoustic communication~\cite{Stojanovic2015} hinders long-range teleoperation of underwater vehicles.  As a result, there is a need for robust and reliable controllers, especially for fully autonomous platforms \cite{Morgan2022}.

Classic approaches to AUV control are done using Proportional–Integral–Derivative controllers (PIDs) \cite{Wan2019, Hedayati2015},
Fuzzy controllers \cite{Khodayari2015, Ghavidel2017} and back-stepping controllers \cite{Ferreira2012}. Each has
its advantages and limitations. While PID controllers are easy to implement and have been intensely studied for multiple decades, they can only control one dimension at a time and cannot exploit the synergies between the different coupled dimensions. In addition, tuning the parameters by hand can be quite time-consuming and challenging.

Although fuzzy controllers do not require a model of the system to work,
they do require expert knowledge to establish the fuzzy rules \cite{Khodayari2015}. Backstepping, if correctly designed, are performant controllers for non-linear systems \cite{Ferreira2012}. However, they are hard to design for underwater systems because of the challenging environment, water currents, hydrodynamics and coupled system dynamics. This makes tuning and finding the right Lyapunov function complex. As they are designed by controlling sub-parts of the system recursively from its model, they are sensitive to modelling errors as well as disturbances. %

Recently, and thanks to more powerful computation units, there is a regain of interest in Model Predictive Controller
(MPC) for AUV control \cite{CARLUCHO2021102726}. MPCs use a predictive model of the controlled system to optimise an action sequence.

One of the main advantages of MPC algorithms is that they can natively perform different tasks. As the optimisation process is performed online, one just needs to change the objective function to achieve a new task. In addition to this, if the cost function is carefully designed, the controller can perform path planning and obstacle avoidance thanks to its feedforward property as well as optimise energy consumption.
However, MPC utilises a linear model, which is not able to model the non-linear dynamics of underwater vehicles \cite{fossen2002,petit2016prediction}.

Another class of model-based controllers is Model Predictive Path Integral (MPPI) controller \cite{williams2017}. MPPI is part of a class of sampled-based MPC algorithms that can be seen as a particle filter operating in the action space.
MPPI provides a gradient-free optimisation alternative, which is very useful for non-linear control problems. Additionally, no constraints are imposed on the model, any modelling technique can be used. Another important characteristic of MPPI is its natural disturbances rejection. These features make MPPI extremely useful for the control of underwater vehicles as it allows flexibility regarding the type of model used, opening the door to many data-driven approaches.

In this work, we investigate the feasibility of MPPI as a suitable control strategy for AUVs.
We developed an MPPI controller utilising a non-linear model, obtained by classical modelling techniques. We then evaluated the proposed controller on a simulation environment \cite{Manhaes2016} and demonstrated its feasibility for application in marine robotics. Additionally, we compared the proposed MPPI controller against two different implementations of a PID controller and compared their performances in a simulated environment. Finally, we present an in-depth analysis of MPPI, studying the impact of the different hyperparameters on the performance of the controlled system. 

This paper is organised as follows:
Section \ref{sec:related} surveys the current literature on the topic.
Section \ref{sec:bg} introduces the dynamical model used in this work. Section \ref{sec:method} presents the MPPI algorithm. Section \ref{sec:exp} highlights the results of the different experiments, and finally,  the conclusions and future work are presented in Section \ref{sec:concl}.

\section{RELATED WORK}
\label{sec:related}

PIDs are one of the most used techniques for control in underwater systems due to their simplicity and low computational load. 
In \cite{Jalving1994} a simple PID for the steering control of an underwater vehicle was proposed. A multiple input multiple output (MIMO) PID block is proposed in \cite{ValenciagaPID} for the control of the Cormoran AUV.  

More recently, a PID controller has been used in combination with sequential observers to control systems subjected to time delays \cite{CarluchoPID}. The sequential observer utilised a data-driven model based on Koopman operators \cite{williams2015data}.

One issue faced by PID controllers is that changes in the working conditions require a re-tuning of the PID gains. To mitigate this issue researchers have explored adaptive techniques.
In \cite{barbalata2015adaptive} an adaptive model-free controller based on PID was developed. The weights of the PID controller were changed online based on the adaptive interaction theory.
While in \cite{CARLUCHO2020280}, a reinforcement learning agent was used to modify the weights of the PID controller. %
However, these techniques still rely on a linear controller for the control of a non-linear platform. Furthermore, it is not straightforward how to introduce constraints in the control system.  

\begin{figure}[ht]
    \centering
    \includegraphics[width=0.45\textwidth]{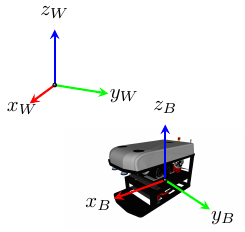}
    \caption{AUV and the two reference frames.}
    \label{fig:model}
\end{figure}

One alternative to PID is MPC controllers, which allow the designers to readily integrate constraints into the cost function. 
In \cite{Gomes2018, Gomes2019} classic MPC was extended to overcome the computational limit of onboard computers. They create a lookup table of known attainable states. \cite{Heshmati2018, Hedayati2015} extend the model to
include a current model. The controller can use the current to produce an energy-efficient solution when the current contributes to the solution. In \cite{Shen2015, Shen2016a, Shen2016b}, the authors investigate the combination of dynamic path planning with non-linear MPC for trajectory tracking. 

While standard MPC controllers are designed for Linear systems, which limits their applicability to underwater vehicles, much work has also been done on non-linear MPC (NMPC) \cite{nmpc}.
Amongst interesting applications, \cite{Nielsen2018} performed docking operations
with a non-linear MPC controller. In \cite{Nikou2020}, it was shown that MPC can be used to operate an AUV mounted with a manipulator.
However, non-linear MPC requires the model to be differentiable and work by linearising the model around points of interest to perform the action sequence optimisation while ours does not.

MPPI was introduced in \cite{Williams2016}. In subsequent work, \cite{williams2017} extended MPPI to work with any kind of forward model.
Different papers have introduced some modifications. In \cite{Hatch2021}, a path planning algorithm's output was used as a terminal cost function. This allows reusing previous path planning computation to improve the performance of MPPI with no additional cost. \cite{Gandhi2021} provided theoretical guarantees on the performance and robustness of MPPI. In \cite{Wagener2019}, the authors re-framed MPPI into a more generic form called Dynamic Mirror Descent-Model Predictive Control, drastically reducing the number of samples at the cost of a small loss in performances.
However, all these works focused on aerial and ground vehicles. It is not clear if MPPI will be effective when operating under the non-linear dynamics of underwater vehicles.

\section{BACKGROUND}
\label{sec:bg}

In this section, we introduce the utilised dynamical model of the AUV \cite{fossen2002}, as well as the basics of model predictive control. 

\subsection{Dynamical Model}

We modelled the AUV utilising the dynamical model introduced in \cite{fossen2002}. We utilise a quaternion representation which helps to avoid singularities. We define two
frames: i) a \emph{world} frame in North-West-Up format and, ii) a \emph{body} frame where the z-axis is aligned with yaw. We show both frames in Fig \ref{fig:model}. 
The dynamics of the vehicle are given by \cite{fossen2002}:
\begin{equation}
    \label{eq:3}
    \begin{split}
        \boldsymbol{(M_{RB} + M_{A}) \dot{\nu} + (C_{RB}(\nu)} &\boldsymbol{+ C_{A}(\nu))\nu}\\
        \boldsymbol{+ D(\nu)\nu + g(m)} &\boldsymbol{= \tau_{C} + \tau_{N}}
    \end{split}
\end{equation}
where $\boldsymbol{M_{RB}}$ is the rigid body mass. $\boldsymbol{M_{A}}$ is the added mass caused by the water displacement.
$\boldsymbol{\nu}$ is the velocity vector expressed in \emph{body}-frame and $\boldsymbol{\dot{\nu}}$ is the acceleration in \emph{body}-frame.
$\boldsymbol{C_{RB}(\nu)}$ are the Coriolis forces. $\boldsymbol{C_{A}(\nu)}$ are the Coriolis forces due to the added mass.
$\boldsymbol{D(\nu)}$ are the damping forces. $\boldsymbol{g(m)}$ are the restoring forces, gravity and buoyancy and $\boldsymbol{m}$ is the pose of the vehicle.
$\boldsymbol{\tau_{C}}$ are the control inputs and $\boldsymbol{\tau_{N}}$ are exterior forces. Grouping together the 
total mass $\boldsymbol{M_{Tot}} = \boldsymbol{(M_{RB} + M_{A})}$, the Coriolis forces $\boldsymbol{C_{Tot}} = \boldsymbol{(C_{RB} + C_{A})}$
we get that:
\begin{equation}
    \label{eq:4}
    \begin{split}
        \boldsymbol{\nu_{t+dt}} &= \boldsymbol{\nu_{t}} + \int_{t}^{t+dt} \boldsymbol{\dot{\nu}_{t}}dt\\
        \boldsymbol{\dot{\nu}} = \boldsymbol{M_{Tot}^{-1}(\tau_{Tot}} & \boldsymbol{- C_{Tot}(\nu)\nu - D(\nu)\nu - g(m))}
    \end{split}
\end{equation}
The kinematics equations are modelled using lie group theory \cite{microLie}. A Lie group is a group that is also a differentiable manifold. The particular Lie group we're using is called $SE(3)$ and is used to represent rigid body motion. The kinematics equation of motion is thus given by:
\begin{equation}
    \label{eq:2}
    \begin{split}
        \boldsymbol{m}_{t+dt} &= \boldsymbol{m}_{t} \oplus \int_{t}^{t+dt}\boldsymbol{\dot{m}}_{t}dt\\
        \boldsymbol{m}_{t+dt} &= \boldsymbol{m}_{t} \oplus \int_{t}^{t+dt}\boldsymbol{Adj}(\boldsymbol{m}_{t})\boldsymbol{\nu}_{t}dt\\
    \end{split}
\end{equation}
Where $\boldsymbol{m}_{t} \in SE(3)$ is the pose of the vehicle, $\boldsymbol{\dot{m}} \in \mathfrak{se}3$ is the velocity in world frame, 
$\boldsymbol{\nu} \in T_{m}SE(3)$ the velocity in \emph{body} frame, $\boldsymbol{Adj}(\boldsymbol{m})$ the adjoint operator that transforms the
velocity from \emph{body} to \emph{world} frame, and $\oplus$ is the right-addition of two $SE(3)$ elements.\\

By setting $\boldsymbol{x} =[\boldsymbol{m}, \boldsymbol{\nu}]$ together with eq \ref{eq:2} and \ref{eq:4}, we can form our trajectory predictor in the form of a recursive function:
\begin{equation}
    \label{eq:5}
    d\boldsymbol{x} = F(\boldsymbol{x}, \boldsymbol{u})
\end{equation}
Where $\boldsymbol{u} \in \mathbb{U} \subseteq \mathbb{R}^{n}$ is the action applied to the system and $n$ is the action dimension.

\subsection{Predictive Model}
\label{subsec:pred_model}
To use the dynamic model presented in the previous subsection with control techniques, such as MPPI, we need to write it as a discrete version. We are interested in obtaining a model of the following prototype:
\begin{equation}
    \label{eq:14}
    \boldsymbol{x}_{t+1} = F(\boldsymbol{x}_t, \boldsymbol{u}_t)
\end{equation}
Recall that the state $x$ is composed of the pose $\boldsymbol{m} \in SE(3)$ and the velocity in body frame $\boldsymbol{\nu} \in T_{m}SE(3)$.
To discrete the system, we assume that the control force $\boldsymbol{\tau_C}$ is constant over a small timestep $\Delta t$. 
This leads to the following update laws for the pose and the velocity:

\begin{equation}
    \label{eq:15}
    \begin{split}
        \boldsymbol{m}_{t+1} &= \boldsymbol{m}_{t} + Adj(\boldsymbol{m}_t) \boldsymbol{\nu}_{t}\\
        \boldsymbol{\nu}_{t+1} &= \boldsymbol{\nu}_{t} + \dot{\boldsymbol{\nu}_{t}}*\Delta t\\
        \dot{\boldsymbol{\nu}_{t}} &= \boldsymbol{M_{Tot}^{-1}(\tau_{Tot}} \boldsymbol{- C_{Tot}(\nu_{t-1})\nu_{t-1}} \\ 
        &\boldsymbol{-  D(\nu_{t-1})\nu_{t-1} - g(m_t))}\\
    \end{split}
\end{equation}

Effectively we implement the update law using \emph{Runge-Kutta} of second order.

\subsection{Model Predictive Control}

MPC is an optimal control technique. It utilises a model of the system to optimise a cost function over a finite horizon window. The minimisation can also take into consideration different system constraints. 
The MPC cost function in a general form can be defined as: 

\begin{equation} \label{eq:J}
\begin{aligned}
\min_{\textbf{u}_{t+ \ell}}   \sum_{\ell = 0}^{\tau} & || \textbf{g}_{t} -  \boldsymbol{\hat{f}(x_{t-1 + \ell}, u_{t-1 + \ell})} ||^2_{{\textbf{Q}}(t)} + || \Delta \textbf{u}_{t+ \ell}||^2_{{\textbf{R}}(t)}  \\  
\textrm{s.t.} \quad & \textbf{u}_{t+ \ell} \in \mathbb{U}^m; \quad \textbf{x}_{t+ \ell} \in \mathbb{X}^n \\ 
\end{aligned}
\end{equation}

\noindent where $x_{ t-1}$ and $u_{ t-1}$ are the state and action taken at time $t -1$ respectively, $\textbf{g}_t $ is the desired position at time $t $, $\tau$ is the horizon window, $ \Delta \textbf{u}_t  $ is the change in control input, and $\textbf{Q}(t )$ and $\textbf{R}(t )$ are weight matrices. The function $\boldsymbol{\hat{f}(\cdot)}$ represents the model used to estimate the future states of the system.
In this simple MPC, the first term of the cost function, presented in Eq. \eqref{eq:J}, takes into consideration the  position error of the system. While the second term seeks to reduce the change between consecutive control efforts, ensuring smooth responses from the motors. This cost function is typical in the MPC literature and is chosen to ensure accurate position tracking and low energy consumption.

\section{METHOD}
\label{sec:method}

In this section, we derive the optimisation process of MPPI and present the algorithm's pseudo-code. 

\subsection{Model Predictive Path Integral Control}

MPPI \cite{williams2017} is a sampled base controller. It is part of the optimal control family, and at its core, it optimises an \emph{user-defined} objective function. The two main principles it is based on are the
\emph{free-energy} and the \emph{Kullback-Leibler divergence}. 
Given a plant model $\boldsymbol{\hat{f}}$ and an objective function $C(\boldsymbol{x_{0:\tau}}): \mathbb{R}^{\tau \times m} \rightarrow \mathbb{R}$, the goal of the controller at every step is to find the action sequence $\boldsymbol{U}^{*} \in \boldsymbol{\mathcal{U}}$ that minimises $C$. First, let's define $\boldsymbol{U} \in \boldsymbol{\mathcal{U}} \subseteq \mathbb{R}^{\tau \times n}$ as the action sequence, $\boldsymbol{u}_t \in \mathbb{U} \subseteq \mathbb{R}^{n}$ an action at time t, and the state at time t $\boldsymbol{x}_t \in \mathbb{X} \subseteq \mathbb{R}^{m}$. While $m$ and $n$ are the state and action dimensions respectively. Finally, the objective function $C$ is defined by:
\begin{equation}
\label{eq:cost_fct}
C(\boldsymbol{x_{0:\tau}}) = \sum_{t=0}^{\tau-1} q(\boldsymbol{x}_t) + \phi(\boldsymbol{x}_{\tau})    
\end{equation}

Where $q(\cdot)$ is called the step-cost and $\phi(.)$ is the final state cost.

In a discrete setting, using a model $\boldsymbol{x}_{t+1} = \boldsymbol{\hat{f}}(\boldsymbol{x}_{t}, \boldsymbol{u}_{t})$. Formally, MPPI solves the following optimisation problem at every timestep:
\begin{equation}
    \label{eq:1}
    \begin{split}
        &\boldsymbol{U^{*}} = \argmin_{\boldsymbol{U}}\sum_{t=0}^{\tau} C(\boldsymbol{x}_{t})\\
    &s.t:\quad \boldsymbol{x_{t+1}} = \hat{f}(\boldsymbol{x}_{t}, \boldsymbol{u}_{t}); \quad \boldsymbol{x} \in \mathbb{X}; \quad \boldsymbol{u} \in \mathbb{U}
    \end{split}
\end{equation}

As mentioned, MPPI is a sample-based controller and the optimisation process takes the following form:
\begin{equation}
    \label{eq:6}
    \boldsymbol{U}^{*} = \argmin_{\boldsymbol{U}} D_{KL}(q^{*}|| q)
\end{equation}

Where $q^{*}$ is the optimal action probability distribution, $q$ the controlled action
distribution and $D_{KL}$ is the Kullback-Leibler divergence. The derivation of $q^{*}$ in
\cite{williams2017} is done with the \emph{free-energy} of the system.

Before introducing the concept of \emph{free-energy} we need to define a few elements. First, let's note that in our application,
the action $\boldsymbol{u}$ corresponds to the control input $\boldsymbol{\tau_{C}}$ of equation \ref{eq:3}.
Next, we define $\boldsymbol{V} = (\boldsymbol{v_{0}}, \boldsymbol{v_{1}}, ..., \boldsymbol{v_{\tau}})$,
a random variable representing an input sequence over a time horizon $\tau$ and where
$\boldsymbol{v_t} \sim \mathcal{N}(\boldsymbol{u_t}, \boldsymbol{\Sigma})$.
This allows us to introduce the uncontrolled action probability distribution $\mathbb{P}$.
That is when $\boldsymbol{u} \equiv 0$. Finally, we define $S: \mathbb{R}^{\tau \times n} \rightarrow \mathbb{R}$ as a cost function over actions. 

The \emph{free-energy} of a control system $(F, S, \boldsymbol{V})$ is defined in \cite{williams2017} as:
\begin{equation}
    \label{eq:7}
    \mathcal{F}(\boldsymbol{V}) = -\lambda \text{log} (\mathbb{E_P} [\text{exp}(-\frac{1}{\lambda} S(\boldsymbol{V})])
\end{equation}
It corresponds to the log of the moment generating function under uncontrolled dynamics.
It is also the equivalent of the value function in path integral control \cite{Theodorou2012}.
The \emph{free-energy} term comes from the Helmholtz free-energy.
It has a similar idea but it is for the controlled system, i.e. the energy is replaced by the \emph{cost-to-go}.
Using Jensen's inequality principle, a change in expectation \cite{williams2017}
and under the assumption that \emph{Radon-Nikodym derivatives} $\frac{d\mathbb{P}}{d\mathbb{Q}}$
and $\frac{d\mathbb{Q}}{d\mathbb{P}}$ exist, we get that:
\begin{equation}
    \label{eq:8}
    \mathcal{F}(\boldsymbol{V}) \leq \mathbb{E_Q}(S(\boldsymbol{V})) + \lambda D_{KL}(\mathbb{Q} || \mathbb{P})
\end{equation}

Meaning that the \emph{free-energy} is bounded from above by the expected \emph{cost-to-go}
under the controlled distribution and a control cost penalising variation of the control
distribution from the uncontrolled distribution, i.e. minimising energy consumption. \cite{williams2017} proved that an
optimal distribution $\mathbb{Q}^{*}$ exists and changes the inequality into an equality.
The probability density function of this optimal distribution takes the form:
\begin{eqnarray}
    \label{eq:9}
    q^{*}(\boldsymbol{V}) = \frac{1}{\eta} \text{exp}(-\frac{1}{\lambda} S(\boldsymbol{V}))p(\boldsymbol{V})
\end{eqnarray}

\begin{algorithm}
    \caption{MPPI algorithm}\label{alg:mppi}
    \KwData{$\hat{\mathbf{f}}$: predictive model;\\ 
    $\mathbf{U} \in \mathcal{U}$ the initial action sequence;\\
    $K$: number of samples;\\
    $\tau$: predictive horizon;\\
    $\boldsymbol{\Sigma}$: Noise generation standard deviation;\\
    $q(\cdot)$: The step cost function;\\
    $\phi(\cdot)$: The terminal cost function;\\
    $\lambda$: The \emph{inverse temperature};}\
    \While{True}{
        $\boldsymbol{x_0} \leftarrow$ get\_robot\_state();\\
        \For{$k \leftarrow$ 0 \KwTo K}{
            $\boldsymbol{\mathcal{E}}^{k} \leftarrow sample(\mathbf{U}, \boldsymbol{\Sigma}, \tau)$; \\
            $\boldsymbol{x} \leftarrow \boldsymbol{x_0}$;\\
            \For{$t \leftarrow 1$ \KwTo $\tau$}{
                $\boldsymbol{x_t} \leftarrow \hat{\mathbf{f}}(\boldsymbol{x}_{t-1}, \boldsymbol{u}_{t-1} + \boldsymbol{\epsilon}^{k}_{t-1}$);\\
                $S(\boldsymbol{\mathcal{E}}^{k}) += q(\boldsymbol{x}_{t}) + \lambda  \boldsymbol{u}_{t-1} \boldsymbol{\Sigma} \boldsymbol{\epsilon}^{k}_{t-1}$;\\
            }
            $S(\boldsymbol{\mathcal{E}}^{k}) += \phi(\boldsymbol{x}_{\tau})$
        }
        $\beta \leftarrow min_{k}[S(\boldsymbol{\mathcal{E}}^{k}]$;\footnotemark \\
        $\eta \leftarrow \sum_{k=0}^{K-1} exp(-\frac{1}{\lambda} (S(\boldsymbol{\mathcal{E}}^{k} - \beta))$;\\
        \For{$k \leftarrow$ \KwTo K-1}{
            $\omega(\boldsymbol{\mathcal{E}}^{k}) \leftarrow \frac{1}{\eta} exp(\frac{1}{\lambda}(S(\boldsymbol{\mathcal{E}}^{k} - \beta))$;\\
        }
        \For{t $\leftarrow$ 0 \KwTo $\tau$}{
            $\boldsymbol{u}_{t} += \sum_{k=0}^{K-1} \omega(\boldsymbol{\mathcal{E}}^{k}) \boldsymbol{\epsilon}_{t}^{k}$;\\
        }
        send\_to\_actuator($\boldsymbol{u}_0$);\\
        \For{$t \leftarrow 1$ \KwTo $\tau-1$}{
            $\boldsymbol{u}_{t-1} \leftarrow \boldsymbol{u}_{t}$;\\
        }
        $\boldsymbol{u}_{\tau-1} \leftarrow init(\boldsymbol{u}_{\tau-1})$;\\
    }
\end{algorithm}
\DecMargin{1em}

\footnotetext{$\beta$ shifts the samples' cost value so that the best sample has a cost of 0. This guarantees that at least one sample will have a non-zero weight.}

Sampling from this distribution is thus equivalent to minimising the \emph{cost-to-go}
and \emph{control-cost}. 
The final step is to find the expression of $\boldsymbol{u}^{*}$ that minimises the KL-divergence. Under the assumption
that the PDF of $\mathbb{Q}$ and $\mathbb{P}$ take the following analytical form:
\begin{align}
    \label{eq:10}
    p(\boldsymbol{V}) &= \prod_{0}^{T-1} Z^{-1} exp(-\frac{1}{2}(v_{t}^{T} \boldsymbol{\Sigma}^{-1} v_{t}))\\
    \label{eq:11}
    q(\boldsymbol{V}) &= \prod_{0}^{T-1} Z^{-1} exp(-\frac{1}{2}((v_{t}- u_{t})^{T} \boldsymbol{\Sigma}^{-1} (v_{t} - u_{t})))
\end{align}

Then \cite{williams2017} showed that:
\begin{equation}
    \label{eq:12}
    \boldsymbol{u}_{t}^{*} = \int_{}^{} q^{*}(\boldsymbol{V}) \boldsymbol{v}_{t} dV
\end{equation}

As one cannot sample from the optimal distribution immediately, \cite{williams2017} derived an importance sampling
scheme. I.E:
\begin{equation}
    \label{eq:13}
    \boldsymbol{u}_{t}^{*} = \mathbb{E_{Q}} [w(\boldsymbol{V})\boldsymbol{v}_t]
\end{equation}
Where $w(\boldsymbol{V})$ is the weight induced by the change from probability distribution $\mathbb{Q}^{*}$ to $\mathbb{Q}$. This also requires to define $\eta$ which is the normalisation term that allows for the weights to represent a probability distribution, i.e $$\sum_{k=0}^{K-1} w(\mathcal{E}^{k}) = 1$$.

For an in-depth derivation of the optimal distribution
$\mathbb{Q}^{*}$, the importance sampling scheme as well as the iterative update law, we invite the reader to look at \cite{williams2017}.

\begin{figure*}[t]
    \centering
    \subfloat[]{%
        \includegraphics[width=0.32\textwidth]{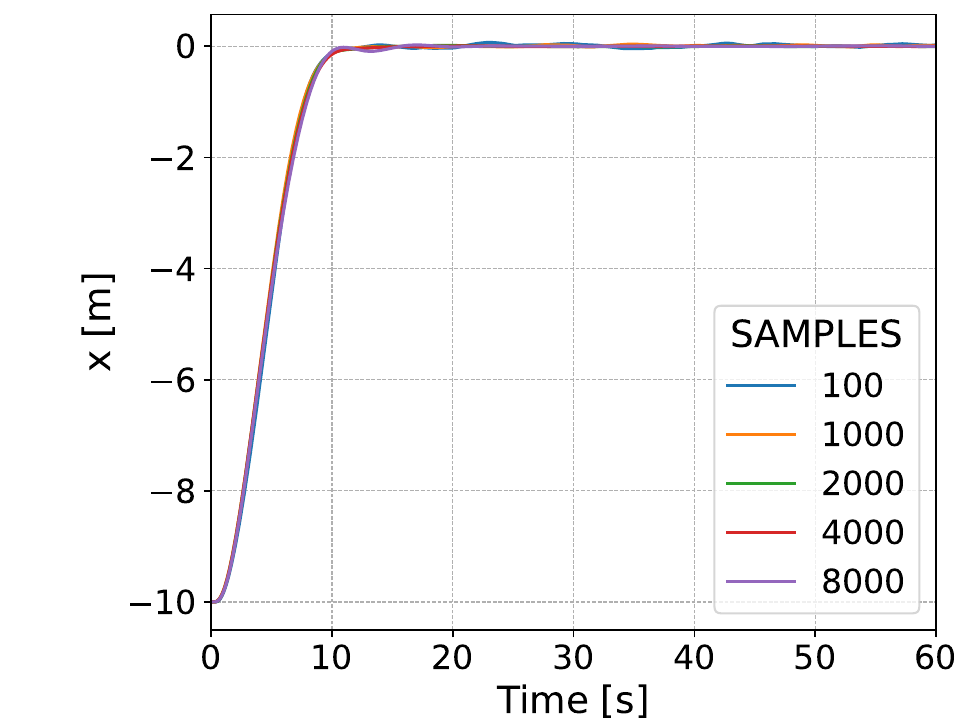}%
        \label{fig:samples_1}%
          } \hfil
    \subfloat[]{%
        \includegraphics[width=0.32\textwidth]{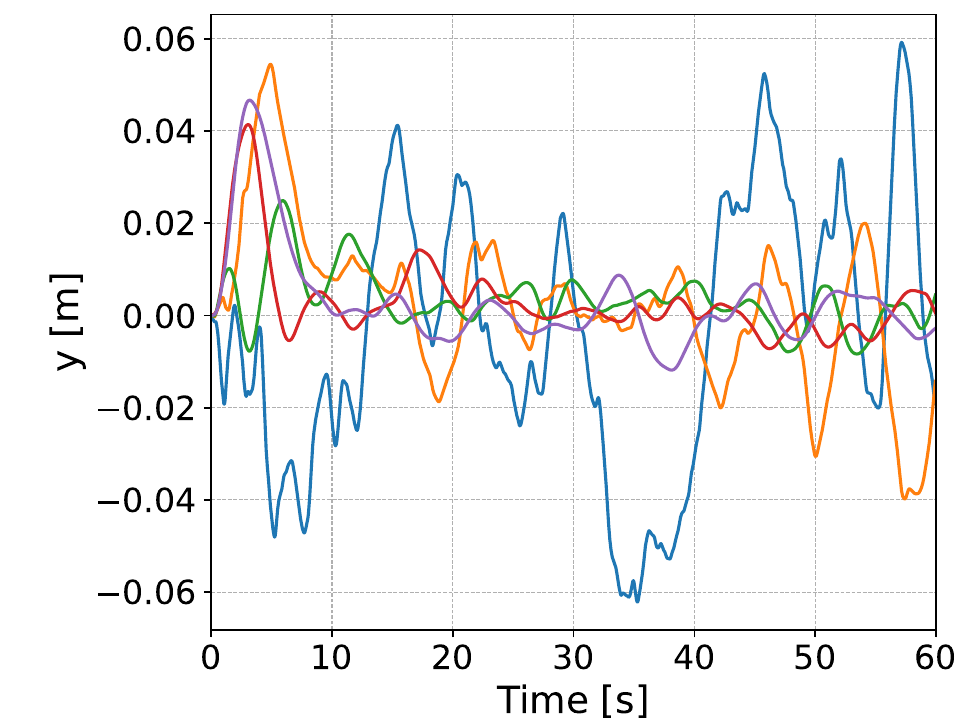}%
        \label{fig:samples_2}} \hfil
            \subfloat[]{%
        \includegraphics[width=0.32\textwidth]{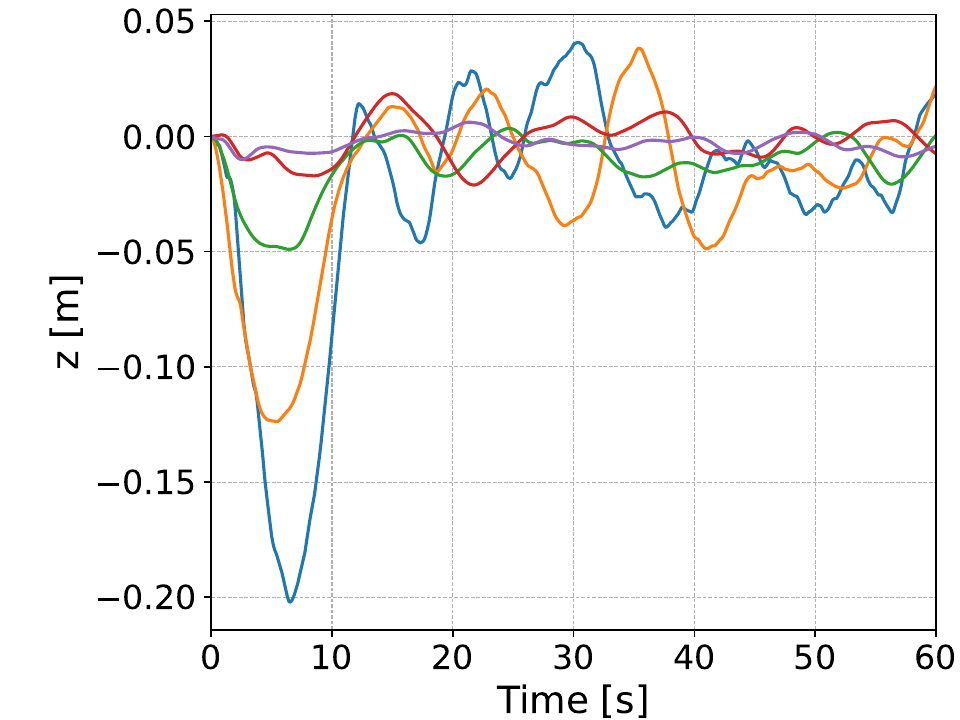}%
        \label{fig:samples_3}%
          } \hfil
    \caption{Analysis of MPPI when varying the number of particles $K$ used for sampling. $\boldsymbol{\Sigma} = 1\%$ of max thrust and $\tau=25$. The target position is $\boldsymbol{x}_{des} = [10,0,0,0]$. a) Error in $x$ axis b) Error in $y$ axis c) Error in $z$ axis.}
    \label{fig:samples}
\end{figure*}

\begin{figure*}[t]
    \centering
    \subfloat[]{%
        \includegraphics[width=0.32\textwidth]{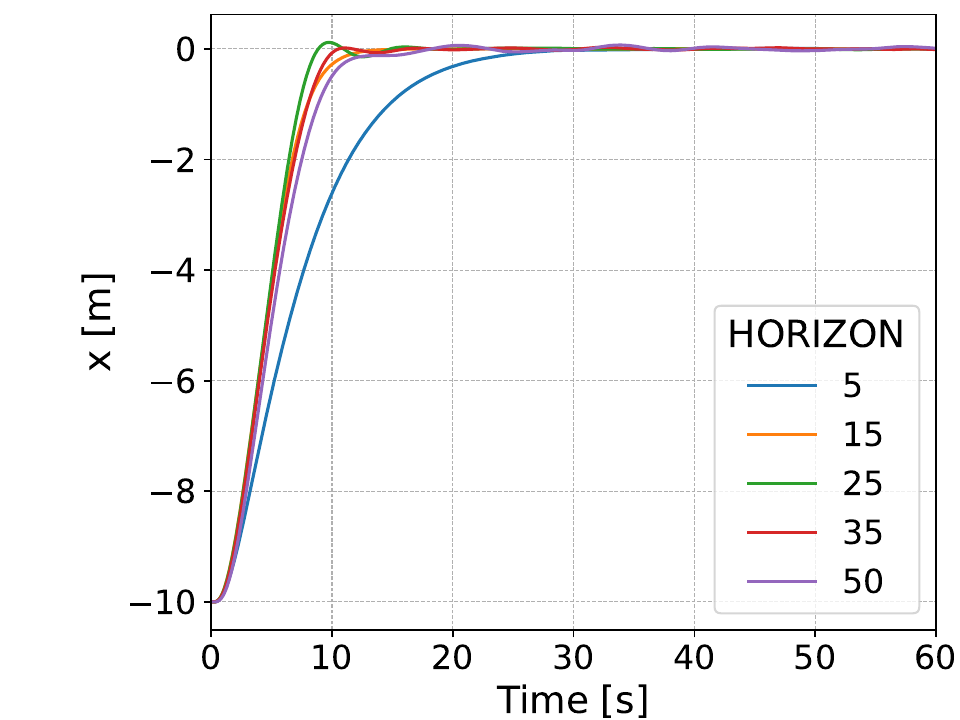}%
        \label{fig:horizon_1}%
          } \hfil
    \subfloat[]{%
        \includegraphics[width=0.32\textwidth]{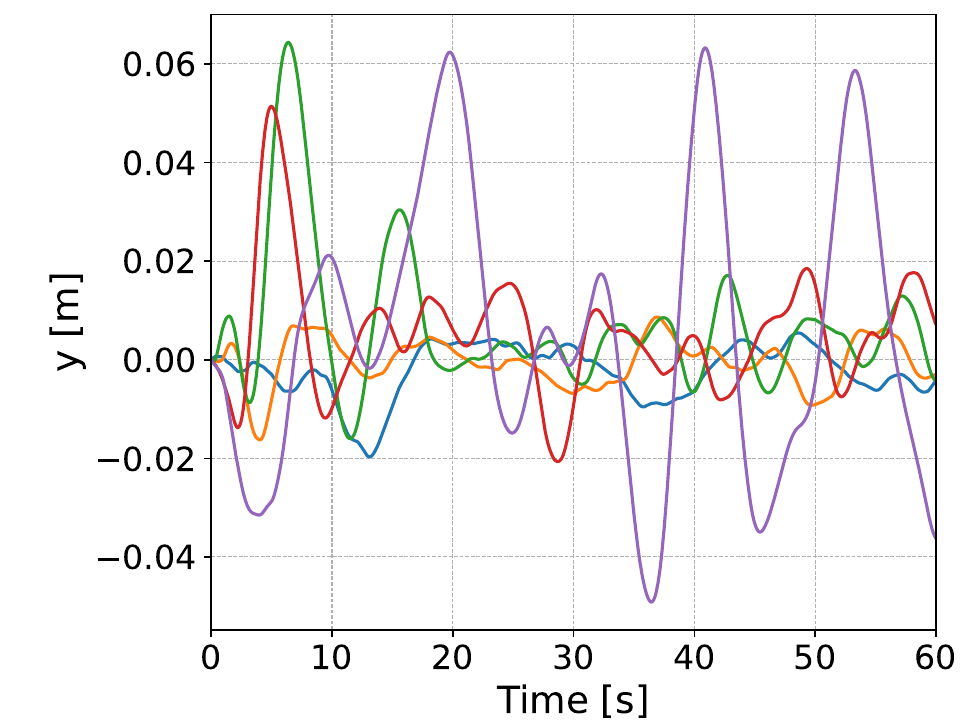}%
        \label{fig:horizon_2}} \hfil
            \subfloat[]{%
        \includegraphics[width=0.32\textwidth]{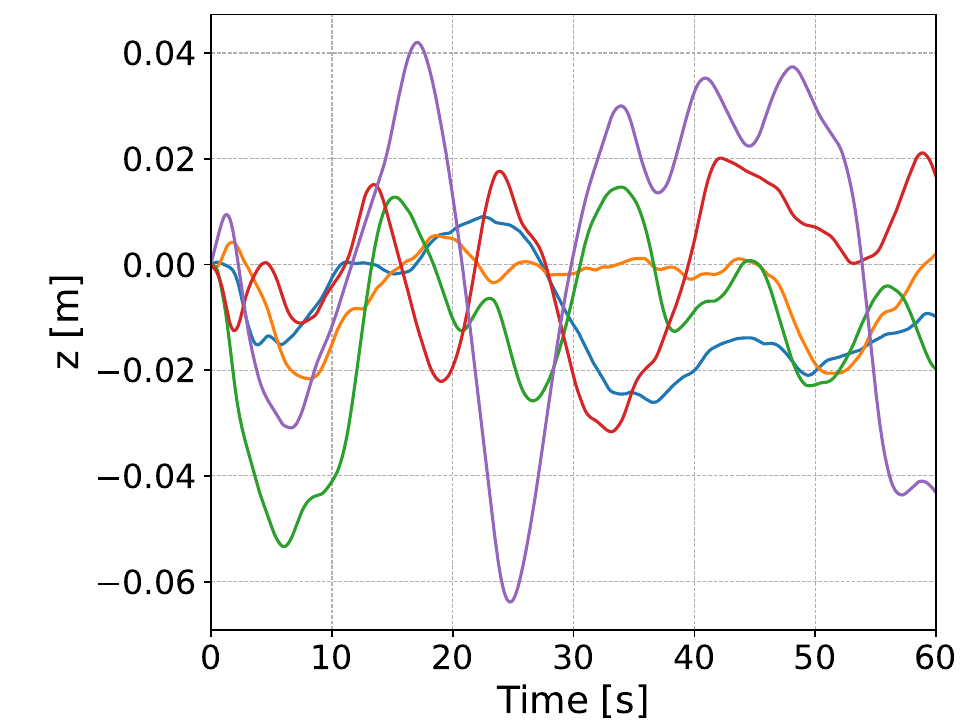}%
        \label{fig:horizon_3}%
          } \hfil
    \caption{Analysis of MPPI when varying the length of the horizon $\tau$. $\boldsymbol{\Sigma}$ is set to 1\% of max thrust and $K=2000$.  The target position is $\boldsymbol{x}_{des} = [10,0,0,0]$.  a) Error in $x$ axis b) Error in $y$ axis c) Error in $z$ axis.}
    \label{fig:horizon}
\end{figure*}

\begin{figure*}[t]
    \centering
    \subfloat[]{%
        \includegraphics[width=0.32\textwidth]{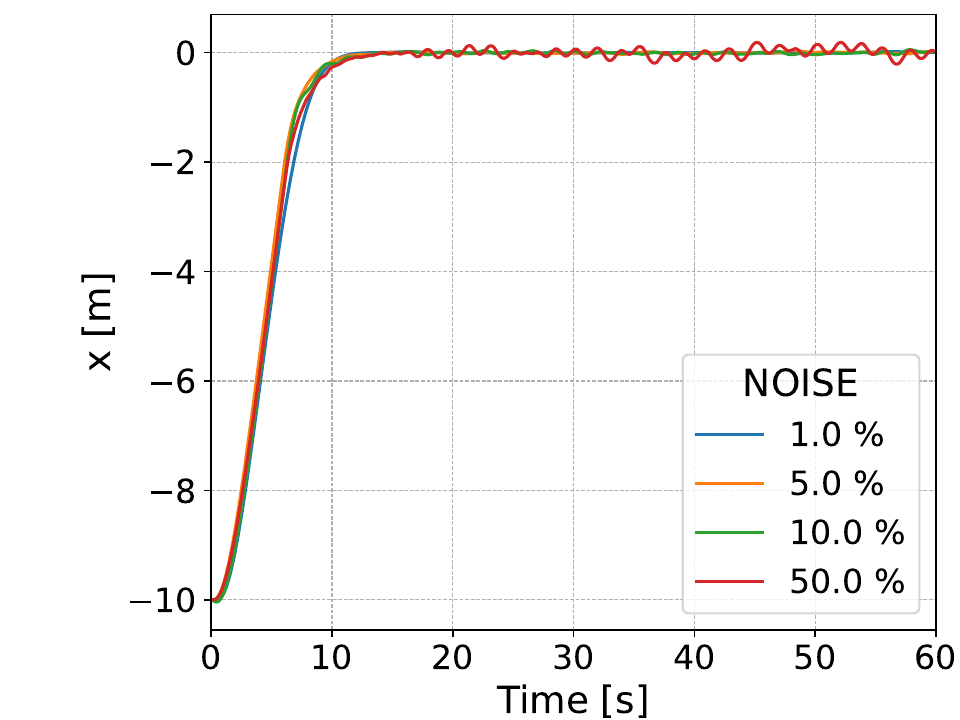}%
        \label{fig:sigma_1}%
          } \hfil
    \subfloat[]{%
        \includegraphics[width=0.32\textwidth]{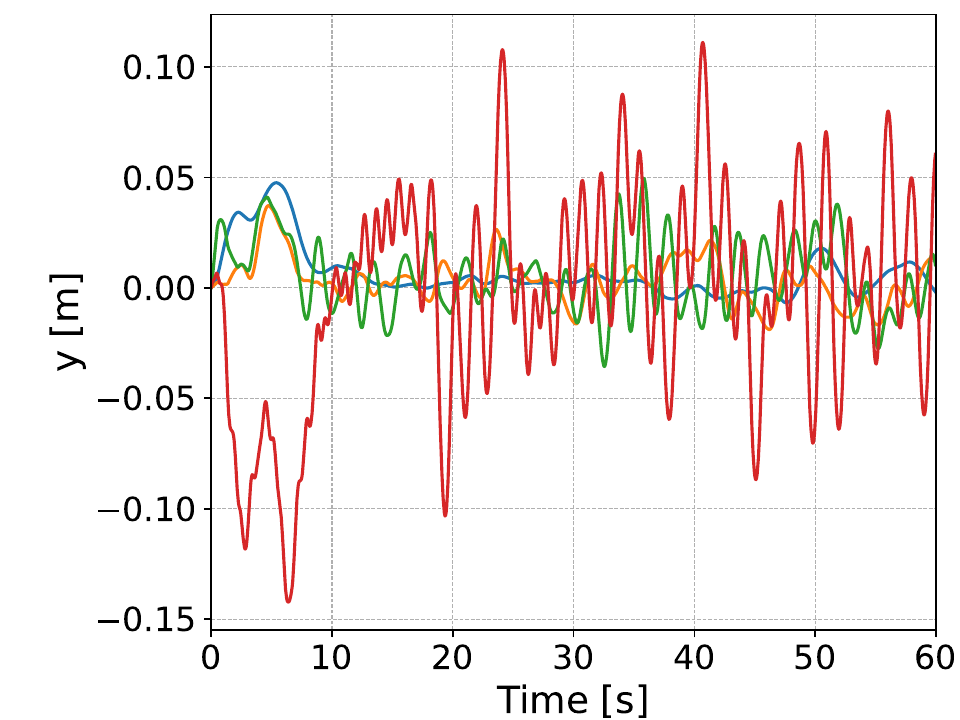}%
        \label{fig:sigma_2}} \hfil
            \subfloat[]{%
        \includegraphics[width=0.32\textwidth]{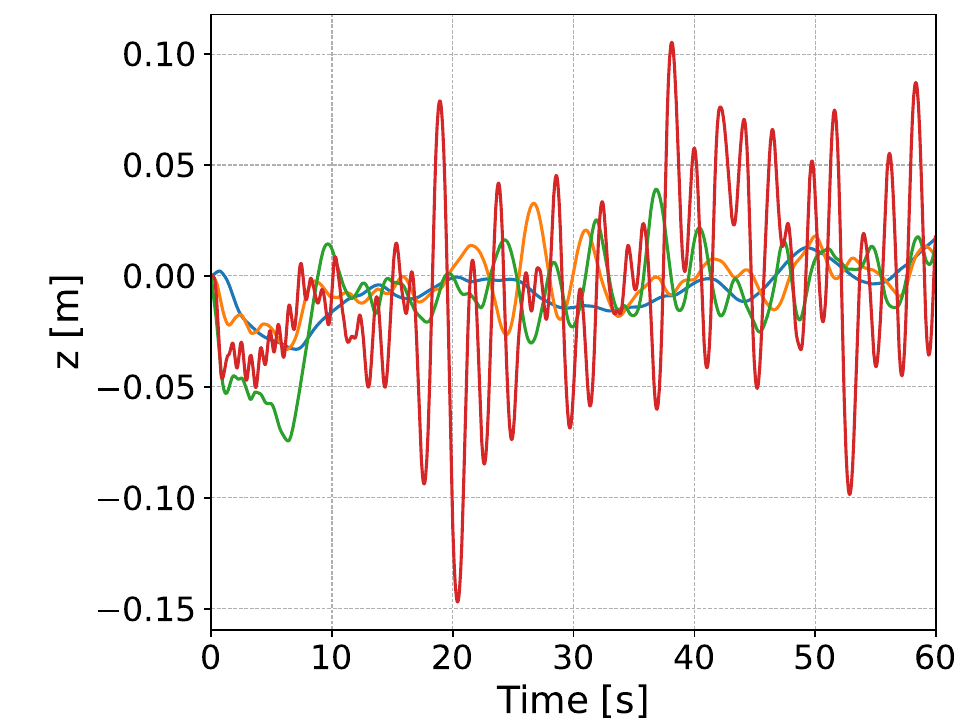}%
        \label{fig:sigma_3}%
          } \hfil
    \caption{Analysis of MPPI when varying the Noise $\boldsymbol{\Sigma}$. The value is expressed in percentage of max thrust. We set $K=2000$ and $\tau=25$. 
    The target position is $\boldsymbol{x}_{des} = [10,0,0,0]$. a) Error in $x$ axis b) Error in $y$ axis c) Error in $z$ axis.}
    \label{fig:sigma}
\end{figure*}

\subsection{Pseudocode}
We now introduce the MPPI pseudocode in Algorithm \ref{alg:mppi}. The algorithm takes as input the cost function and several hyperparameters. The main hyperparameters for the controller are $\lambda$ the inverse temperature, which controls how many samples contribute to the decision-making. $\tau$ the predictive horizon, controlling the feed-forward part. $K$, the number of samples and $\boldsymbol{\Sigma}$ the noise generation standard deviation. In the main loop, the MPPI algorithm first samples an action sequence for each sample, and then generates a rollout with the model for each of the samples. Then it computes the samples' weight and updates the action sequence accordingly. Note that the samples are independent of one another and thus the rollouts can be executed concurrently.

\begin{figure*}[t]
    \centering
    \subfloat[]{%
        \includegraphics[width=0.32\textwidth]{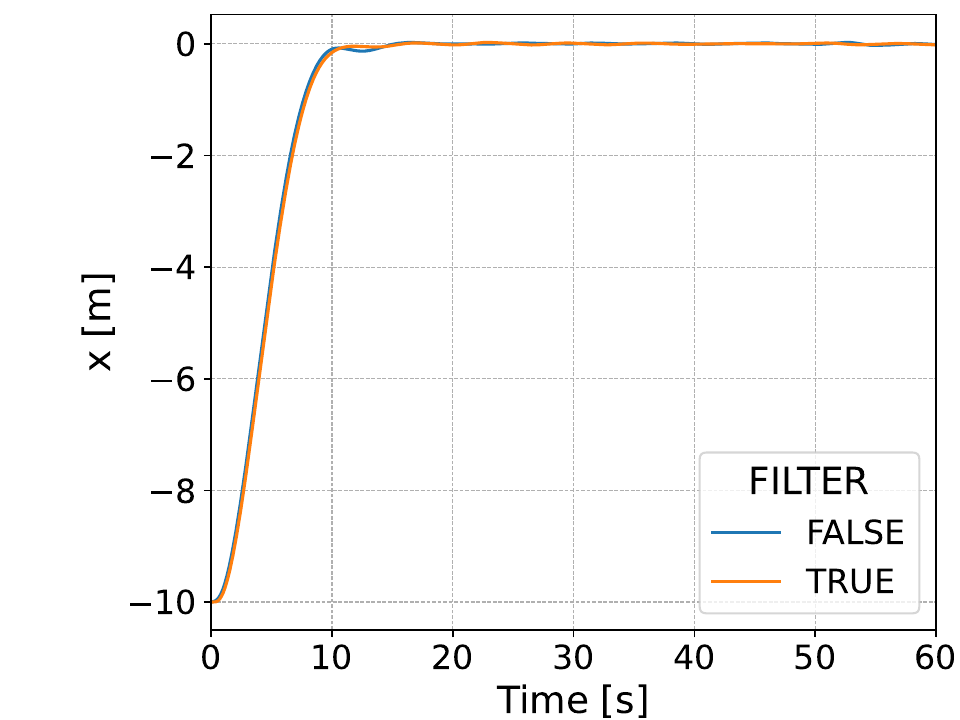}%
        \label{fig:filter_seq_1}%
          } \hfil
    \subfloat[]{%
        \includegraphics[width=0.32\textwidth]{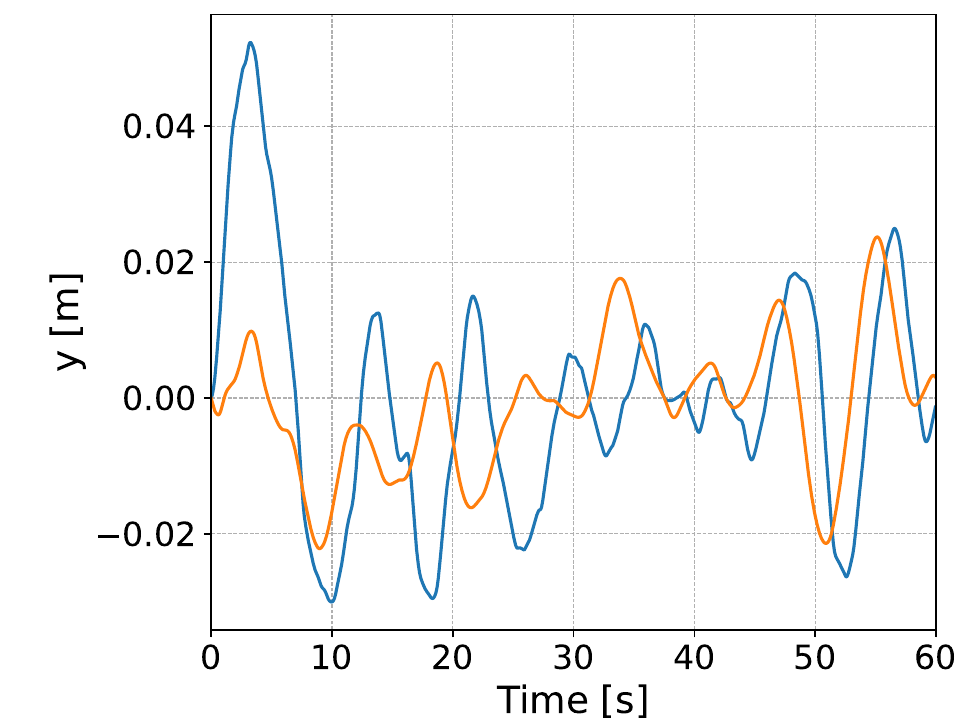}%
        \label{fig:filter_seq_2}} \hfil
    \subfloat[]{%
        \includegraphics[width=0.32\textwidth]{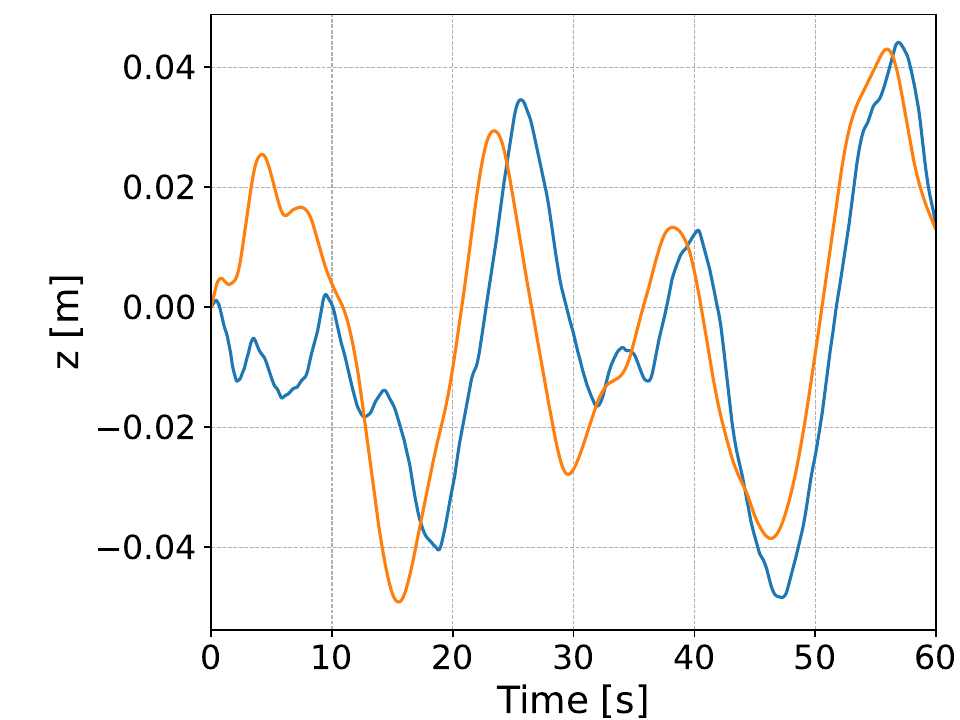}%
        \label{fig:filter_seq_3}} 
    \caption{Analysis of filtering the action sequence with the Savitzky-Golay filter on the position error. a) The error on the $x$-axis. b) The error on the $y$-axis. c) The error on the $z$-axis.}
    \label{fig:filter_seq}
\end{figure*}

\begin{figure*}[t]
    \centering
    \subfloat[]{%
        \includegraphics[width=0.32\textwidth]{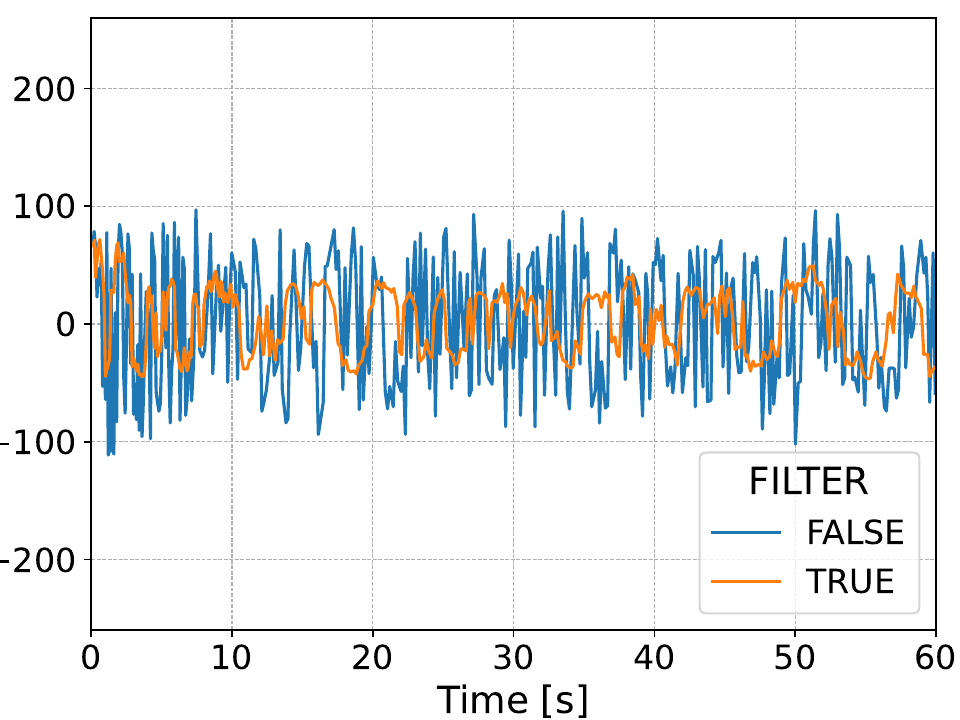}%
        \label{fig:filter_seq_thruster_1}%
          } \hfil
    \subfloat[]{%
        \includegraphics[width=0.32\textwidth]{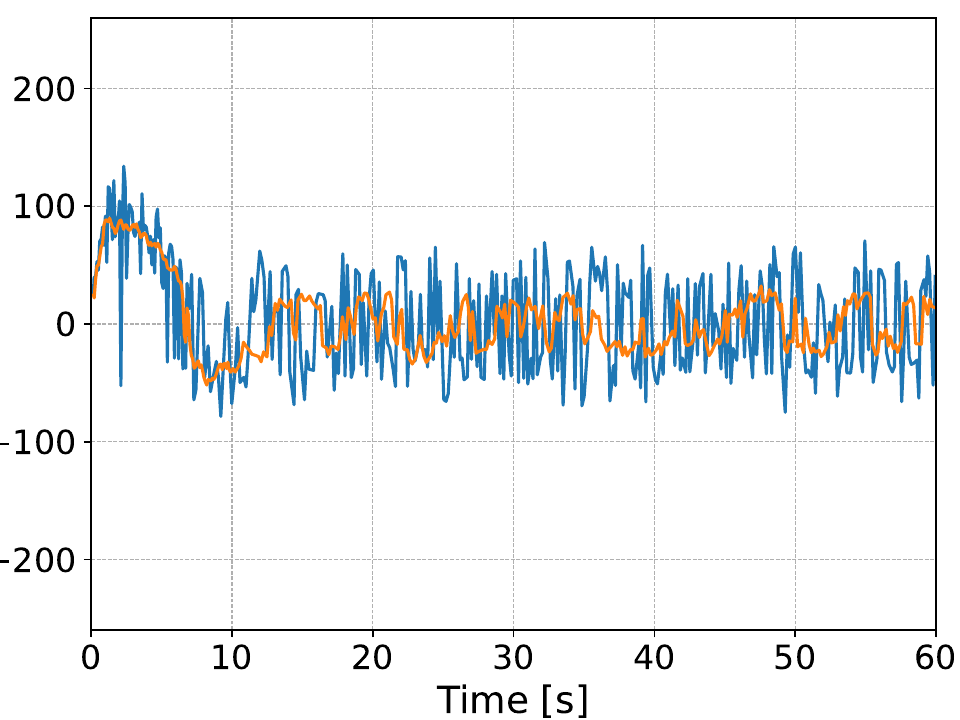}%
        \label{fig:filter_seq_thruster_2}} 
    \subfloat[]{%
        \includegraphics[width=0.32\textwidth]{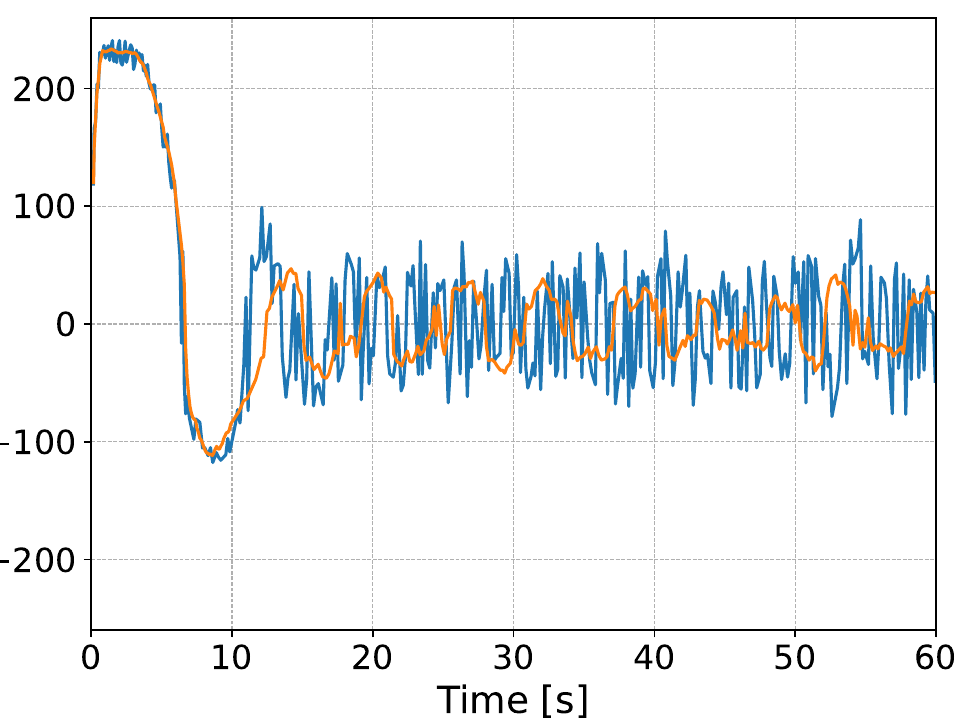}%
        \label{fig:filter_seq_thruster_3}} 
    \caption{Effect of the Savitzky-Golay filter on different thrusters input. a) Thruster 0 b) Thruster 2 c) Thruster 4.}
    \label{fig:filter_seq_thruster}
\end{figure*}

\begin{figure}[t]
    \centering
    \subfloat[]{%
        \includegraphics[width=0.24\textwidth]{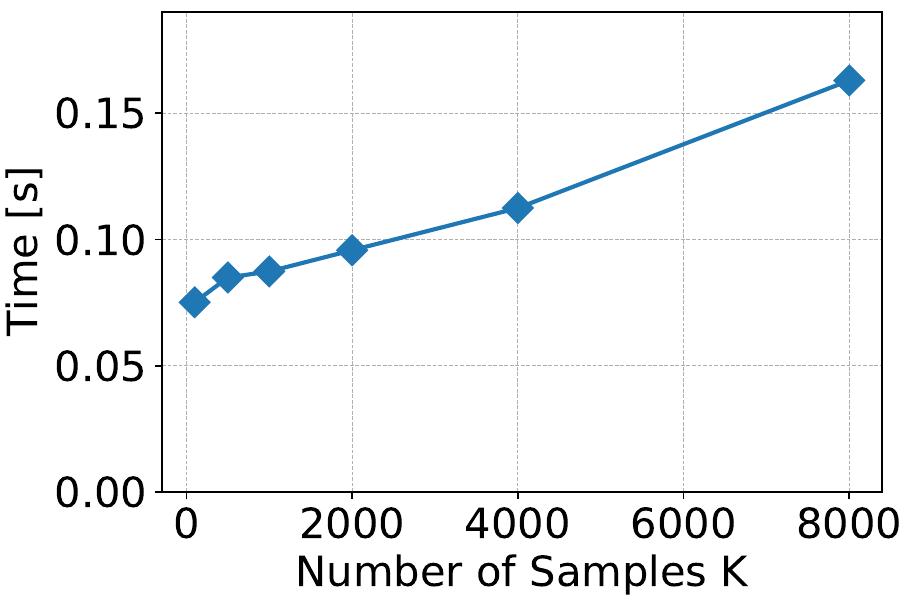}%
        \label{fig:time_samples}%
          } \hfil
    \subfloat[]{%
        \includegraphics[width=0.24\textwidth]{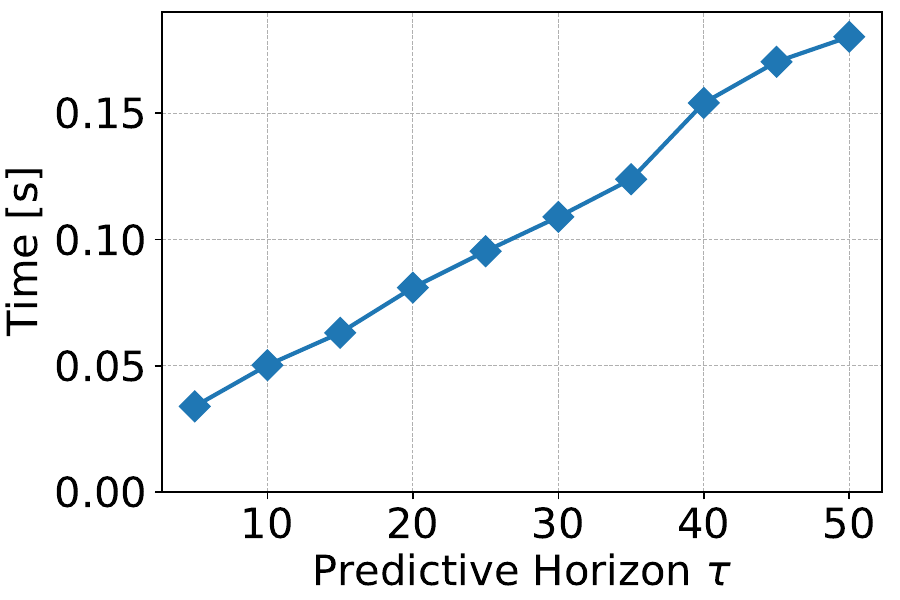}%
        \label{fig:time_tau}} 
    \caption{Timing of MPPI for various parameters. a) Variation of the number of samples $K$, when $\tau$=25  b) Variation of the predictive horizon $\tau$, when $K=2000$.}
    \label{fig:timing}
\end{figure}

\begin{figure}[t]
\centering
  \includegraphics[width=.8\linewidth]{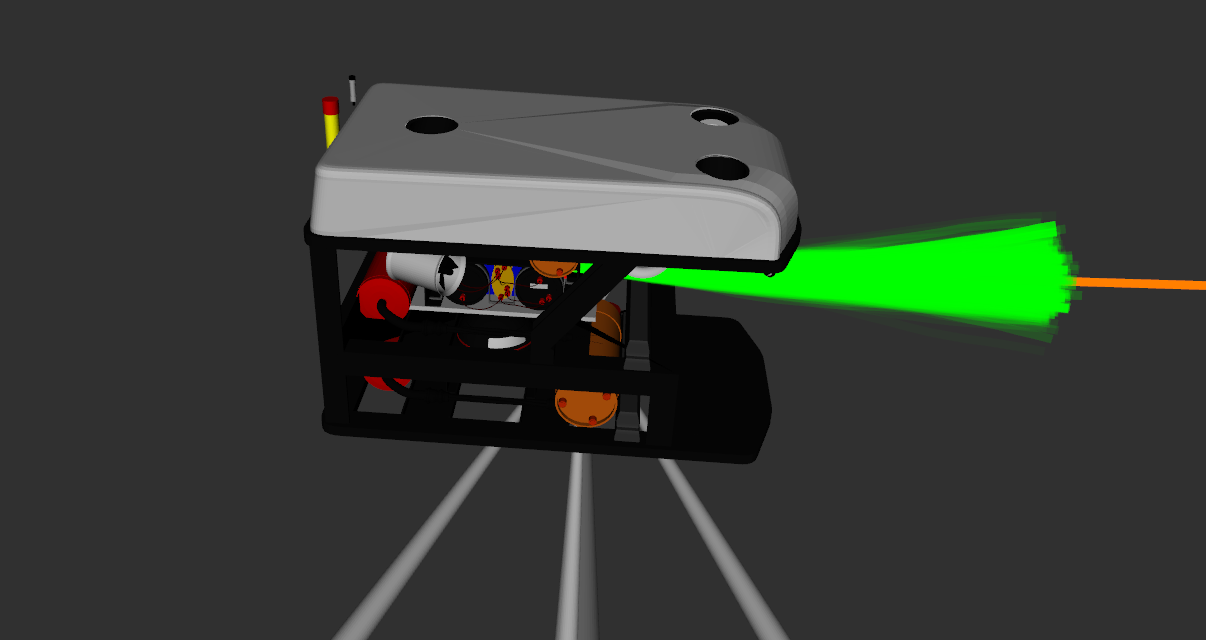}
\caption{Convergence of MPPI samples.}
\label{fig:convergence}
\end{figure}

\begin{figure*}[t]
    \centering
    \subfloat[]{%
        \includegraphics[width=0.248\textwidth]{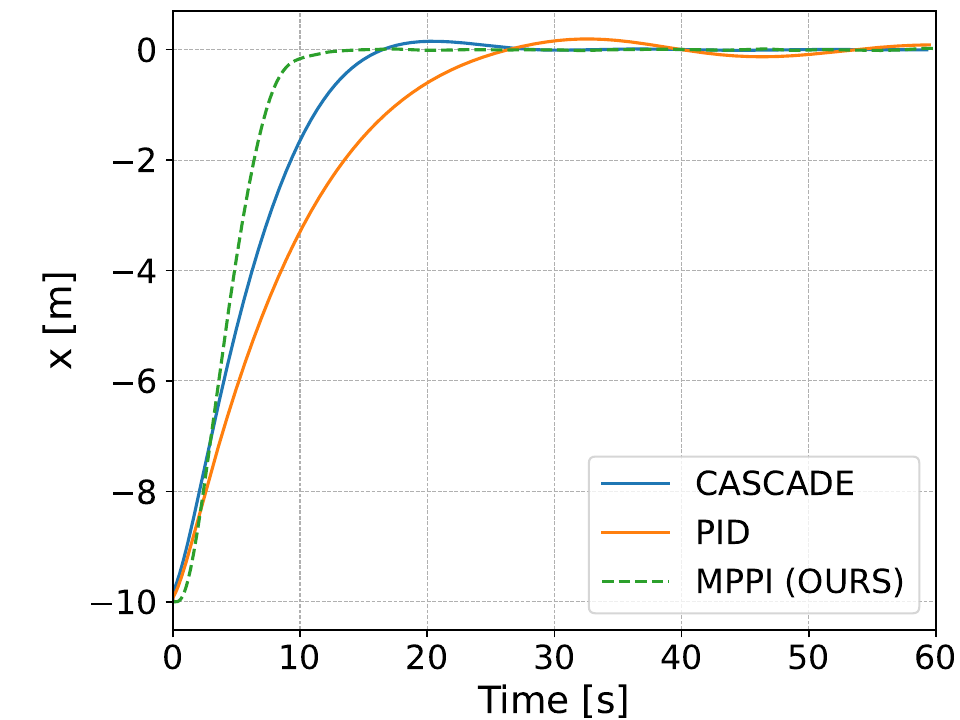}%
        \label{fig:comparison_neutral_1}%
        } \hfil
    \subfloat[]{%
        \includegraphics[width=0.248\textwidth]{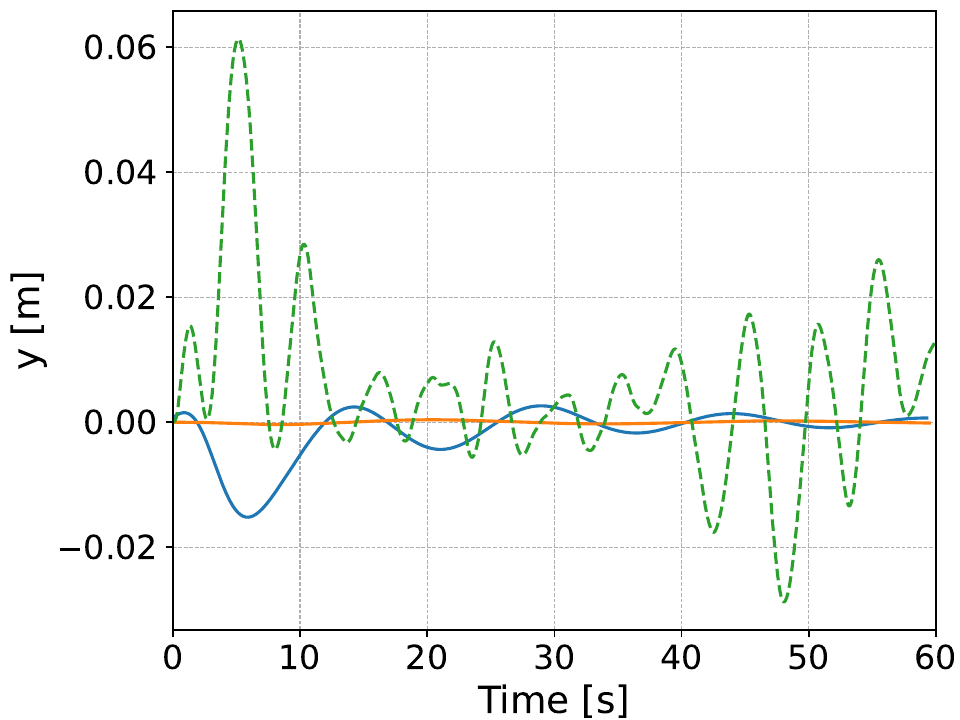}%
        \label{fig:comparison_neutral_2}%
        }
    \subfloat[]{%
        \includegraphics[width=0.248\textwidth]{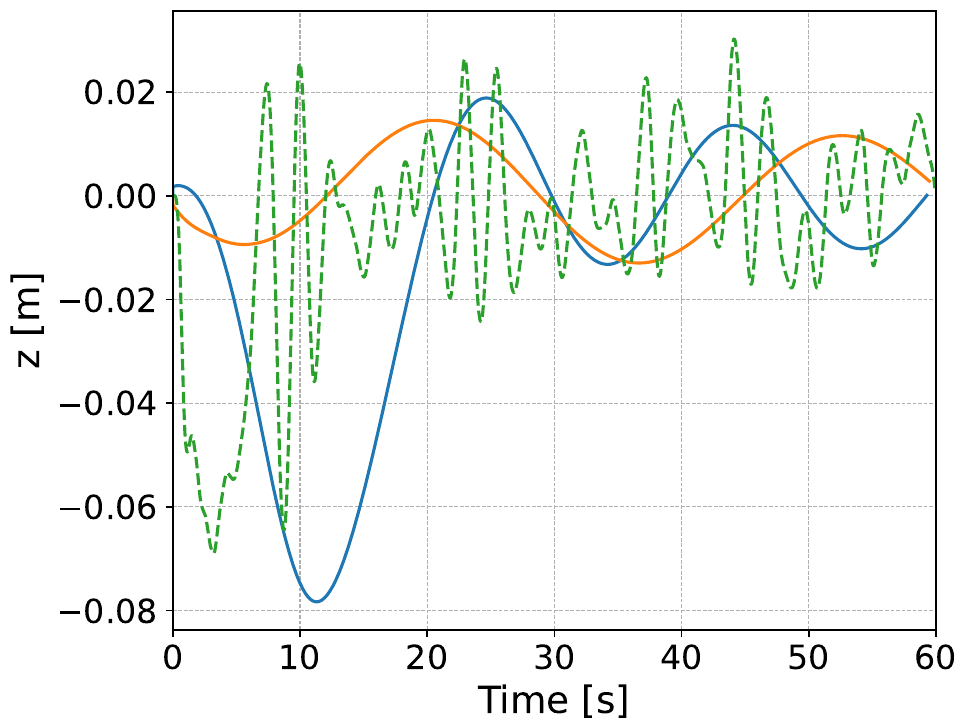}%
        \label{fig:comparison_neutral_3}%
        }
    \subfloat[]{%
        \includegraphics[width=0.248\textwidth]{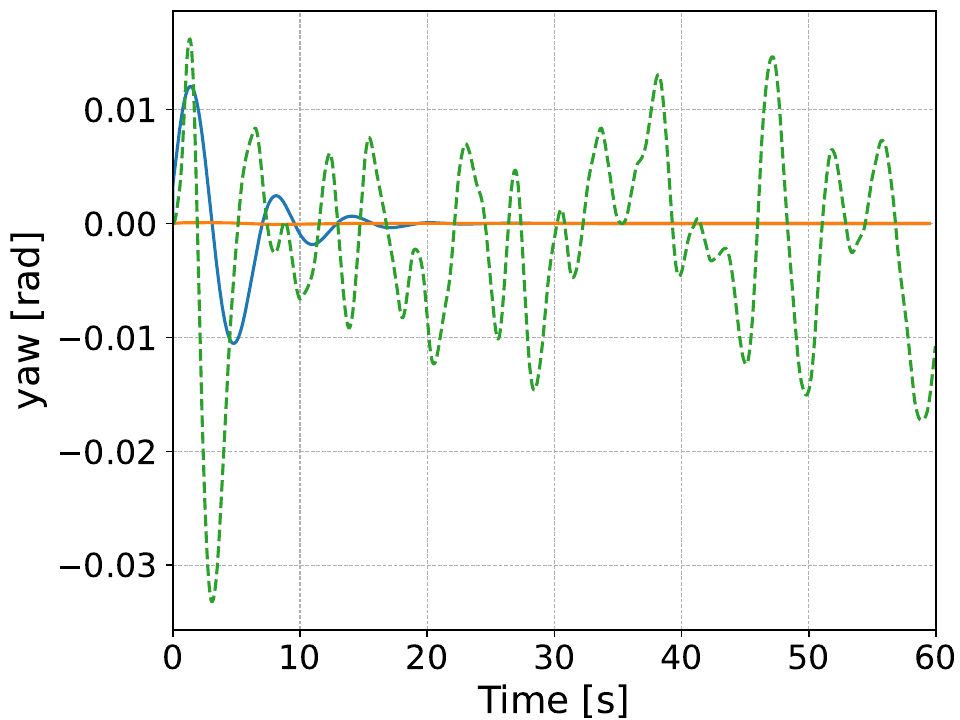}%
        \label{fig:comparison_neutral_4}%
        }
    \caption{Comparison between MPPI, PID and  CASCADE PID on the forward task with a neutrally buoyant vehicle. a) The error on the $x$-axis. b) The error on the $y$-axis. c) The error on the $z$-axis. d) The error on yaw.}
    \label{fig:comparison_neutral}
\end{figure*}

\begin{figure*}[t]
    \centering
    \subfloat[]{%
        \includegraphics[width=0.248\textwidth]{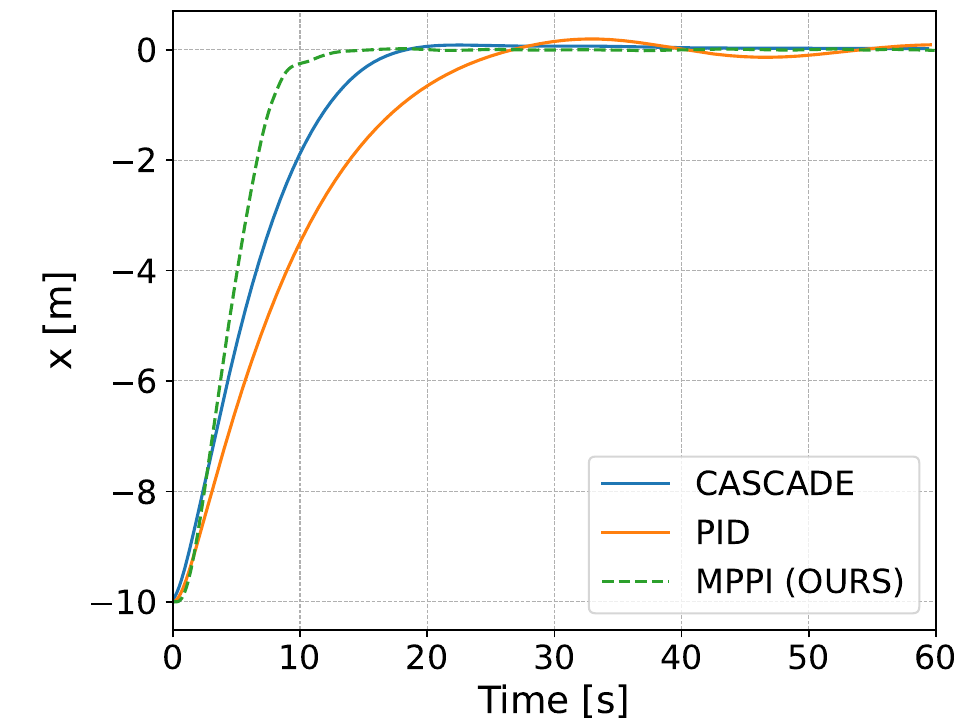}%
        \label{fig:comparison_neg_1}%
          } \hfil
    \subfloat[]{%
        \includegraphics[width=0.248\textwidth]{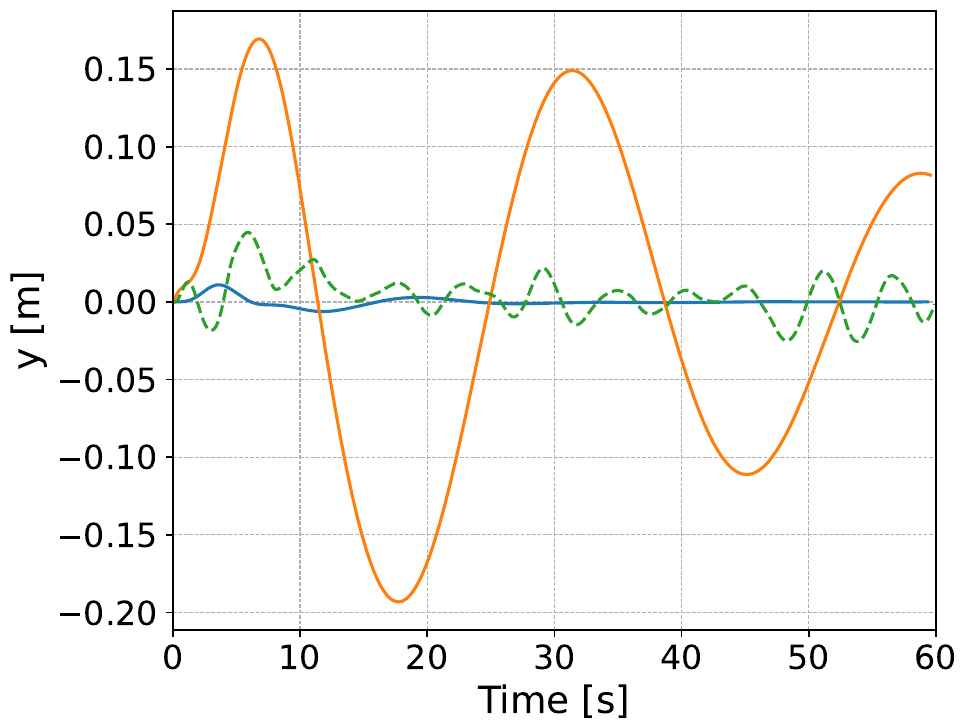}%
        \label{fig:comparison_neg_2}%
        }
    \subfloat[]{%
        \includegraphics[width=0.248\textwidth]{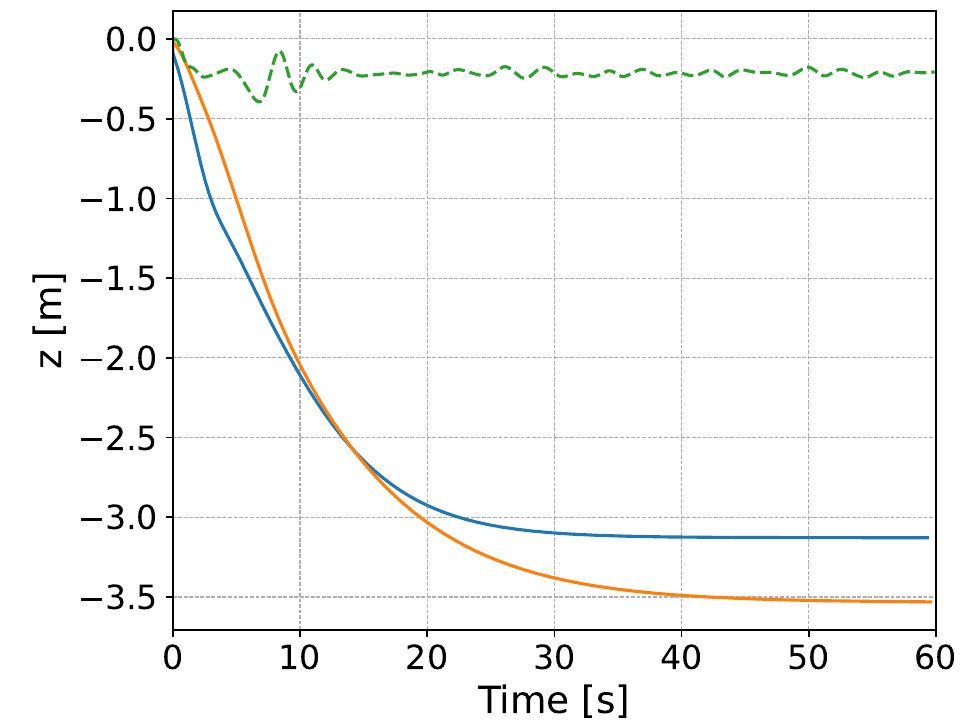}%
        \label{fig:comparison_neg_3}%
        }
    \subfloat[]{%
        \includegraphics[width=0.248\textwidth]{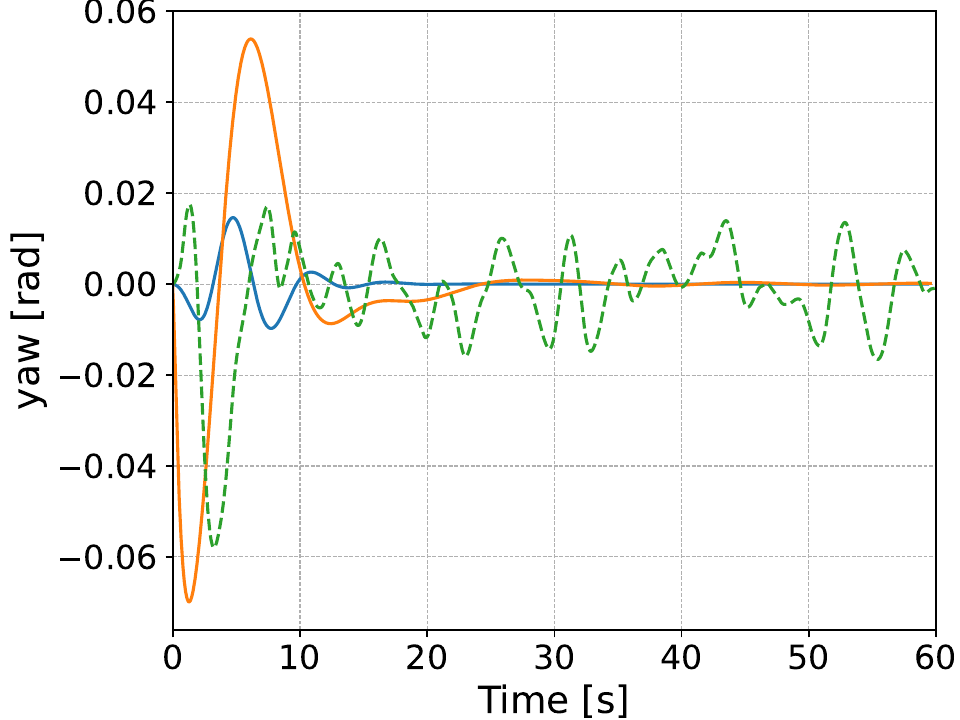}%
        \label{fig:comparison_neg_4}
        }
    \caption{Comparison between MPPI, PID and  CASCADE PID on the forward task with a negatively buoyant vehicle. a) The error on the x-axis. b) The error on the y-axis. c) The error on the z-axis. d) The error on yaw.}
    \label{fig:comparison_neg}
\end{figure*}

\section{EXPERIMENTS}
\label{sec:exp}

In this section, we introduce the experiments performed to evaluate our proposed controller. All experiments in this section were performed with the UUV simulator \cite{Manhaes2016}. We utilised the RexROV2 AUV\footnote{\url{https://github.com/uuvsimulator/rexrov2}} as a robotic platform. The code to perform all of our experiments is available on GitHub\footnote{\url{https://github.com/NicolayP/mppi-rexrov2}}.
We performed three sets of experiments. First, we investigate the impact of the different hyperparameters on the performance of the MPPI controller. Second, we provide a set of experiments to compare the performance of  MPPI against a PID. Finally, we demonstrate the advanced capabilities of MPPI by evaluating it on more challenging tasks that require simultaneous control of multiple degrees of freedom while at the same time performing obstacle avoidance.

We focus on the tasks of waypoint navigation. For each task, the reference consists of one static point indicating the final desired pose and velocity of the vehicle.
The step cost function used for MPPI takes the form of a scaled $L_2$ norm:
\begin{equation*}
   L(\boldsymbol{x}) = L_2(\boldsymbol{x}, \boldsymbol{x}_{des}, \boldsymbol{Q}) = 
   ||(\boldsymbol{x}-\boldsymbol{x}_{des}) ||_{\boldsymbol{Q}}
\end{equation*}

\noindent with $\boldsymbol{x}$ being the composition of the pose and the velocity of the vehicle, $\boldsymbol{x}_{des}$ being the desired pose and velocity. $\boldsymbol{Q}$ is a scaling matrix that allows to change the importance of every axis. %
The error between the quaternions, used to define the rotation, $q$ and $q_{des}$ is computed by $angle((q^{-1}*q_{des}))$ where \emph{angle} extracts the angle from the \emph{angle-axis} representation of the quaternion. The terminal cost $\phi(\cdot)$ is set to $L(\boldsymbol{x})$ as well.

For all the experiments the matrix $\boldsymbol{Q}$ is set to $\boldsymbol{Q} = diag(10, 10, 10, 100, 10, 10, 10, 10, 10, 10)$
Where the first three elements match the position, the fourth one is for the angular term and the last 6 ones match the velocities. The \emph{inverse temperature} $\lambda$, is set to 0.06.
The selection of the number of samples $K$, the predictive horizon $\tau$, and the noise generation $\boldsymbol{\Sigma}$ is detailed in the next subsection. For the initial action sequence $\mathbf{U}$, we utilised a vector of zeros of the corresponding length. 
    
\subsection{MPPI Analysis}
\label{sec:mpppi_analysis}

In this subsection, we evaluate the impact of the different hyperparameters on the controller's performance. We evaluate three main parameters: i) the number of samples $K$, ii) the predictive horizon $\tau$, and iii) the noise $\Sigma$. Finally, we also show the changes in performance when applying a filter on the action sequence to provide a smoother control signal.

\paragraph{Samples} In the first experiment we vary the number of samples $K$ utilised by the controller. To evaluate the changes in performance, we task the AUV with reaching a certain point 10m forward. We perform the same experiment while varying the number of samples $K$. 
The resulting error on \emph{x, y, z}-axis is given in Figure \ref{fig:samples}.
As expected, the number of samples increases the precision of the controller. This is similar to particle filters increasing the number of samples, which allows for a wider exploration of the search space. We show the sample convergence in Figure \ref{fig:convergence}.
Increasing the number of samples, however, is at the cost of computational power. However, our implementation utilises TensorFlow, which allows us to parallelize the rollout for each sample on the GPU, which leads to faster computation. The limit then depends on the number of cores the GPU has.

Based on Figures \ref{fig:samples} and \ref{fig:time_samples} it can be seen that $K=2000$ achieves the best compromise between performances and computational time. Therefore, for the following experiments, we set the number of samples $K$ to $2000$.

\paragraph{Horizon} The next experiment consists in varying the predictive horizon. The error on \emph{x, y, z}-axis is reported in Figure \ref{fig:horizon}. The figure shows that as the horizon increases, the precision of the controller increases. However, above 35 prediction steps, the controller performance decreases. We consider two potential origins of this effect. The first one is that, as the predictive horizon increases, so does the search space. If we do not increase the number of samples, the search space is not explored efficiently, thus leading to poor results. The second potential cause might result from samples where the first steps are very jittery but the last elements randomly converge towards the goal. This leads to those samples having a higher weight and therefore having more impact on the decision-making. We selected $\tau=25$ as the optimal value.

\paragraph{Noise} The third hyperparameter we investigate is the noise standard deviation $\boldsymbol{\Sigma}$. The matrix is a diagonal matrix where every entry indicates the noise generated along that axis. 

The error on \emph{x, y, z}-axis is reported in Figure \ref{fig:sigma}. As expected the higher the noise the more jittery the controller. However, if the noise is too small, the response time and natural noise rejection of the controller will be impacted. We found good performances when the noise is set to 1\% of the max input.

\paragraph{Hyperparameter Selection} In addition to the performance, we are also interested in developing a controller that can run in real-time. To this end, we performed an evaluation of the run-time when varying the hyperparameters that have a higher impact on performance. 
For these experiments, the noise standard deviation $\boldsymbol{\Sigma}$ is set to $1\%$ of the max thrust and $\lambda = 0.06$. The choice of $\lambda$ is done according to \cite{Williams2019}, where they claim that MPPI is \emph{healthy} when $\eta$, the normalisation term, is between $1-10\%$ of the number of samples.  The obtained computational time of the controller for the different parameters can be found in Figure \ref{fig:timing}. 
First, we fixed the horizon to $\tau = 25$, and varied the number of particles. As can be seen in Figure \ref{fig:time_samples}, $K=2000$ allows us to achieve a $10$ Hz run-time. We then varied the horizon $\tau$ while maintaining the number of particles at $2000$ (Figure \ref{fig:time_tau}). Keeping the horizon $N$ below 30 samples, allows us to achieve a frequency of $10$ Hz. 
Therefore, we identify $\tau = 25$ and $K=2000$, as suitable hyperparameters for our problem.

\paragraph{Filtering} Finally we also show the impact of adding a filter on the action sequence. 
For these experiments we utilise a Savitzky-Golay filter \cite{Savgol_filter}. 
The results obtained with the filter can be seen in Figures \ref{fig:filter_seq} and \ref{fig:filter_seq_thruster}. The first noticeable effect is to reduce the natural jittering of MPPI and thus increase its precision. The second and more interesting effect is to filter out the control signal. This avoids high-frequency control signals that might be intractable and/or damage real-world thrusters.

\subsection{Comparison with PID}
{
    \def\OldComma{,}
    \catcode`\,=13
    \def,{
      \ifmmode
        \OldComma\discretionary{}{}{}
      \else
        \OldComma
      \fi
    }
}
\label{sec:comparison}
In this subsection, we evaluate the proposed MPPI controller, with the hyperparameters as defined in the previous subsection, against two different types of PIDs: a single PID and a  
cascade PID \cite{CascadePID}. For the single PID, we used the following parameters $K_p $ $=$ ${\rm diag}(250, \allowbreak 250, \allowbreak 250, \allowbreak 800, \allowbreak 800, 800)$, 
$K_i$ $=$ $\allowbreak {\rm diag}(100,\allowbreak 100, \allowbreak 100, 300, \allowbreak 300, \allowbreak 300)$ and 
$K_d $ $=$ $ {\rm diag}(1950, \allowbreak 1950, \allowbreak 1950, 1000, \allowbreak 1000, \allowbreak 1000)$.
While for the cascade PID, we used 
$K^p_p $ $=$ $ {\rm diag}(10, 10, 10, \allowbreak 35,  \allowbreak 35,  \allowbreak 35)$, $K^p_i = {\rm diag}(1.5, 1.5, 1.5, \allowbreak 10, 10, 10)$ and 
$K^p_d$ $=$ $ {\rm diag}(35,  \allowbreak 35,  \allowbreak 35, \allowbreak 25, 25, 25)$ for the position controller, and 
$K^v_p $ $=$ $ {\rm diag}(30, 30, 30, \allowbreak 50, 50, 50)$, 
$K^v_i$ $=$ ${\rm diag}(15, 15, \allowbreak 15, 40, \allowbreak 40, 40)$ and 
$K^v_d $ $=$ $ \allowbreak {\rm diag}(25, \allowbreak 25, 25, 30, 30, 30)$ for the velocity controller.

\begin{figure*}[t]
    \centering
    \subfloat[]{%
        \includegraphics[width=0.45\textwidth, trim={7.5cm 3cm 4.5cm 4cm},clip]{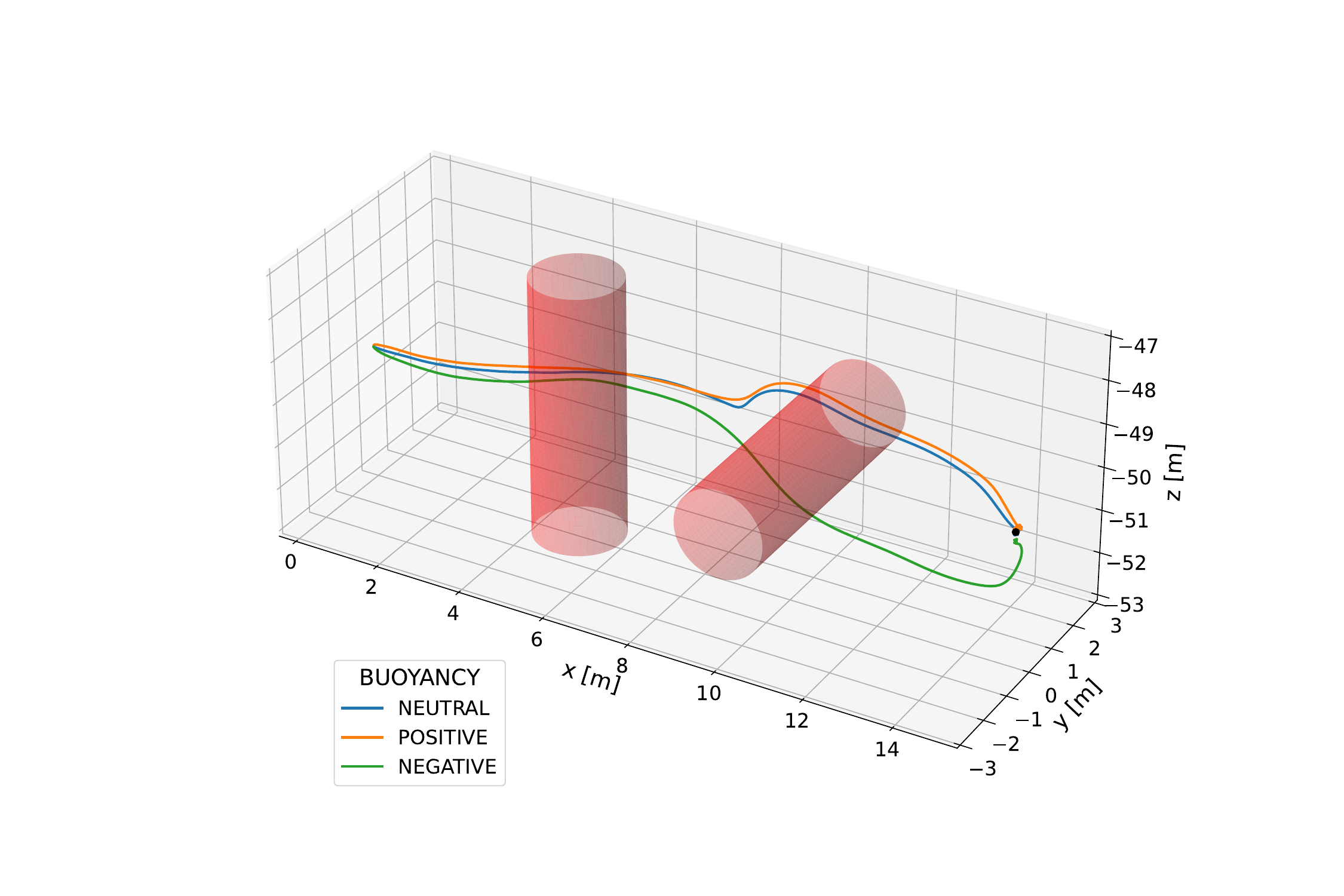}%
        \label{fig:ob1}%
          } \hfil
    \subfloat[]{%
        \includegraphics[width=0.45\textwidth]{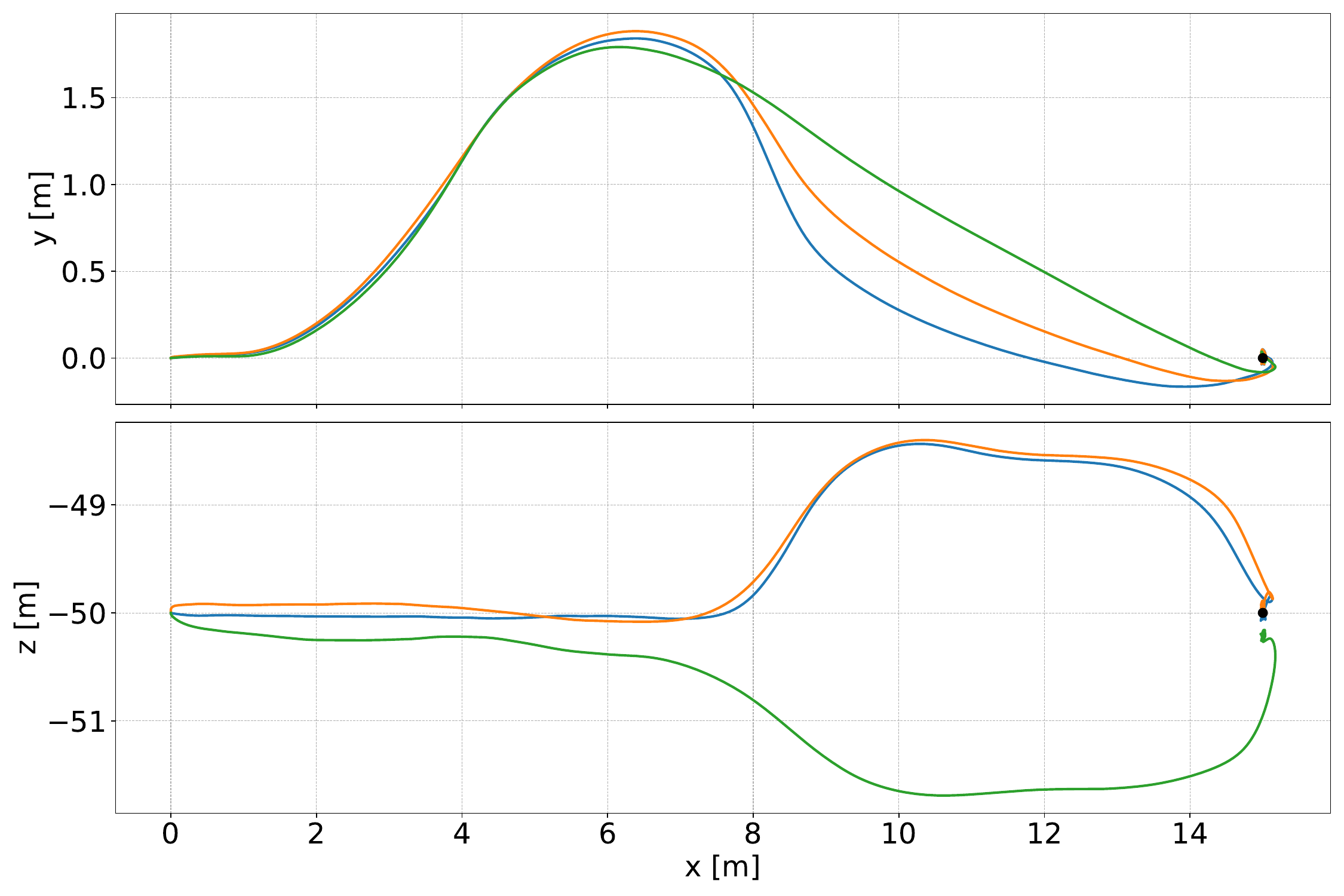}%
        \label{fig:ob2}}
    \caption{Resulting trajectory when utilising MPPI to avoid obstacles. We evaluated three different dynamic configurations: i) Neutral ii) Positive buoyant and iii) Negative buoyant. The goal is located on the black dot, while the red cylinders represent the obstacles. a) Resulting 3D trajectory b) Projection of the trajectory, on the $xy$-plane (top), and the $xz$-plane (bottom).}
    \label{fig:obs}
\end{figure*}

We first evaluate MPPI and compare it against the PIDs when performing a simple task. We provide all controllers with the task of moving forward 10 meters. The results obtained in this task can be seen in Figure \ref{fig:comparison_neutral}. We can see that MPPI has a faster response time. It reaches the desired position in less time and without overshoot. On the other hand, both PID controllers are slower and present a small overshoot in the x-axis. However, all controllers are able to maintain their position in the $y$ and $z$ axis, while at the same time presenting almost no variation in the yaw angle.

We next evaluate the controllers' ability to adapt to changes in the dynamics of the AUV. We maintain the same hyperparameters for all controllers, but we modified the AUV. To achieve this we add a virtual mass of 100kg to the vehicle. This effectively makes the AUV negatively buoyant, and it also changes the rigid body kinematics of the vehicle.
Other than this, all three of the controllers remain unchanged throughout the experiments and we keep the same goal as the previous task. The obtained trajectories for the modified configuration are shown in Figure \ref{fig:comparison_neg}.
It can be clearly seen how MPPI outperforms both PIDs. While the dynamics in the x-axis for all controllers remain similar, a big difference can be seen in the z-axis. In this axis, both PIDs present significant steady-state errors, while the MPPI controller is able to maintain the vehicle in the required position on the z-axis. We argue that the predictive capabilities of MPPI allow it to optimally control the vehicle, even after a change in the dynamics occurs. 

\subsection{Obstacle avoidance experiments}

In this subsection, we present additional experiments to highlight the capability of MPPI to perform a complex control task. The experiment seeks to highlight the natural obstacle avoidance feature of MPPI, a feature lacking in other approaches such as PIDs. 
To evaluate this capability, we place two cylinders on the path to the goal. We assume that the controller has a perception system that allows it to determine the exact position of the obstacles, which allows us to incorporate them into the cost function. 
The controller then checks for collisions between each sampled trajectory and the obstacles. 
If an individual sample collides with the obstacle, we assign to it a \emph{infinite} cost value, thus effectively discarding that sample from the decision-making. This provides a low computational solution that can be readily applied to our original cost function. 

Additionally, and to further showcase the ability of MPPI to adapt to changes in the dynamics, we evaluate the controller under three different operative conditions: i) Neutrally buoyant configuration, ii) positive buoyant configuration and, iii) negative buoyant configuration. The hyperparameters are the same as those selected in Section \ref{sec:mpppi_analysis}, which were tuned for the neutral buoyant configuration of the AUV and stayed constant for all the experiments. The change in buoyancy is then performed by adding virtual weights to the AUV for the negatively buoyant configuration or by virtually increasing the volume for the positively buoyant configuration. The negative configuration is the same as in \ref{sec:comparison}. The positively buoyant configuration has a virtual volume increase that brings the restoring forces to 250N.

The obtained results can be seen in Figures \ref{fig:ob1} and \ref{fig:ob2}. It can be clearly seen how the MPPI controller is able to direct the AUV to the desired position, while at the same time avoiding the obstacles. Furthermore, MPPI is able to modify the trajectories to account for the different AUV dynamics. For example, for the negative buoyant vehicle, the controller takes the lower path under the second obstacle. While for the neutral and positive buoyant configuration, it takes the path above the obstacle. We believe these results demonstrate the flexibility and effectiveness of MPPI as a control system for AUVs.

\section{CONCLUSION}
\label{sec:concl}

In this article, we evaluated the feasibility of MPPI as a control technique for an autonomous underwater vehicle. We developed an MPPI controller utilising a non-linear model of the AUV. We presented several experiments in simulation showing the advantages of our method.
We evaluated the performance of the MPPI when varying different hyperparameters, and obtained the optimal configuration which would allow for real-time operation while maintaining optimal performance. 
Furthermore, we provided a comparison against a PID controller which clearly shows the advantage of our method.  
Finally, we present results in a simulation environment that shows how MPPI can be used to solve more complex tasks, such as obstacle avoidance, by providing small changes to the cost function. 

In future work, we propose to test the proposed controller in a real underwater platform.  Additionally, we believe that the MPPI architecture can be combined with data-driven modelling techniques to improve the performance of the system and increase its adaptability to dynamic changes.

\bibliographystyle{IEEEtran}
\bibliography{myBib}

\end{document}